\documentclass[journal,twoside,web]{ieeecolor}
\usepackage{tmi}
\usepackage{cite}
\usepackage{amsmath,amssymb,amsfonts}
\usepackage{algorithmic}
\usepackage{graphicx}
\usepackage{booktabs}
\usepackage{multirow}
\usepackage{graphicx}
\newcommand{\sigm}{\operatorname{sigmoid}}
\usepackage{amsmath,amssymb,mathtools}
\usepackage{algorithm}
\usepackage{algorithmic}
\usepackage{bm}
\usepackage{xcolor}

\newcommand{\DWT}{\mathrm{DWT}}
\newcommand{\iDWT}{\mathrm{iDWT}}
\newcommand{\Proj}{\Pi}

\floatname{algorithm}{Algorithm}

\usepackage{subcaption} % 仅此，无需全局 \captionsetup
\renewcommand\thesubfigure{\Alph{subfigure}}

\usepackage{subcaption}  % 推荐
\usepackage{array} % 很重要！提供 m{..} 等列型
\newcolumntype{C}[1]{>{\centering\arraybackslash}m{#1}} % 定宽+水平居中+垂直居中

\usepackage{textcomp}
\def\BibTeX{{\rm B\kern-.05em{\sc i\kern-.025em b}\kern-.08em
    T\kern-.1667em\lower.7ex\hbox{E}\kern-.125emX}}
\usepackage{makecell}

\begin{document}
\title{Layer-wise Noise Guided Selective Wavelet Reconstruction for Robust Medical Image Segmentation}
\author{Yuting Lu, Ziliang Wang, Weixin Xu, Wei Zhang,Yongqiang Zhao , Yang Yu and Xiaohong Zhang 
\thanks{Yuting Lu, Weixin Xu, Wei Zhang and Xiaohong Zhang are with the Big Data and Software College, Chongqing University, Chongqing 40031, China.  }
\thanks{Yongqiang Zhao, is with the Office of Scientific Research, Peking University.}
\thanks{Ziliang Wang is with the Key Lab of High Confidence Software Technology, Ministry of Education (Peking University), 
and the School of Computer Science, Peking University, Beijing 100871, China.}
\thanks{Yang Yu is with Department of Cardiology and Institute of Vascular Medicine, Peking University Third Hospital, Beijing 100191, China   }
\thanks{Corresponding author: Ziliang Wang. Email: wangziliang@pku.edu.cn.}
% \thanks{T. C. Author is with the Electrical Engineering Department,
% University of Colorado, Boulder, CO 80309 USA, on leave from the National
% Research Institute for Metals, Tsukuba, Japan (e-mail: author@nrim.go.jp).}
}

\maketitle

\begin{abstract}
Clinical deployment requires segmentation models to stay stable under distribution shifts and perturbations. The mainstream solution is adversarial training (AT) to improve robustness; 
however, AT often brings a clean--robustness trade-off and high training/tuning cost, which limits scalability and maintainability in medical imaging.
We propose \emph{Layer-wise Noise-Guided Selective Wavelet Reconstruction (LNG-SWR)}. During training, we inject small, zero-mean noise at multiple layers to learn a frequency-bias prior that steers representations away from noise-sensitive directions. 
We then apply prior-guided selective wavelet reconstruction on the input/feature branch to achieve frequency adaptation: suppress noise-sensitive bands, enhance directional structures and shape cues, and stabilize boundary responses while maintaining spectral consistency. 
The framework is backbone-agnostic and adds low additional inference overhead. It can serve as a plug-in enhancement to AT and also improves robustness without AT.
On CT and ultrasound datasets, under a unified protocol with PGD-$L_{\infty}/L_{2}$ and SSAH, LNG-SWR delivers  consistent gains on clean Dice/IoU and significantly reduces the performance drop under strong attacks; combining LNG-SWR with AT yields additive gains. 
When combined with adversarial training, robustness improves further without sacrificing clean accuracy, indicating an engineering-friendly and scalable path to robust segmentation.
These results indicate that LNG-SWR provides a simple, effective, and engineering-friendly path to robust medical image segmentation in both adversarial and standard training regimes.

\end{abstract}


\section{Introduction}
\label{sec:introduction}
\begin{figure}[t]
  \centering
  % Row 1
  \begin{subfigure}[t]{0.488\linewidth}
    \centering
    \includegraphics[width=\linewidth]{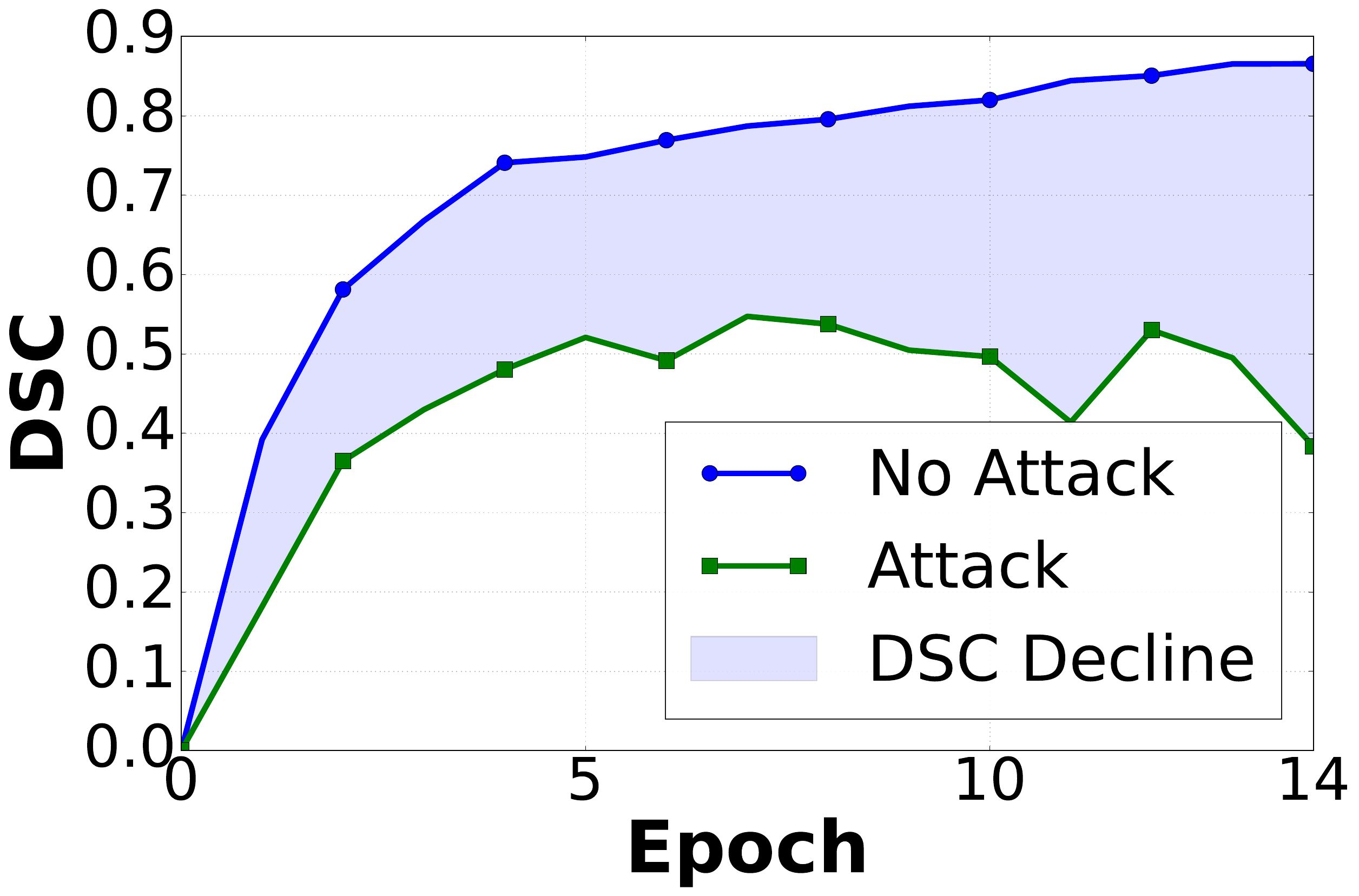}
    \subcaption{}\label{fig:a}
  \end{subfigure}\hspace{0.008\linewidth}
  \begin{subfigure}[t]{0.488\linewidth}
    \centering
    \includegraphics[width=\linewidth]{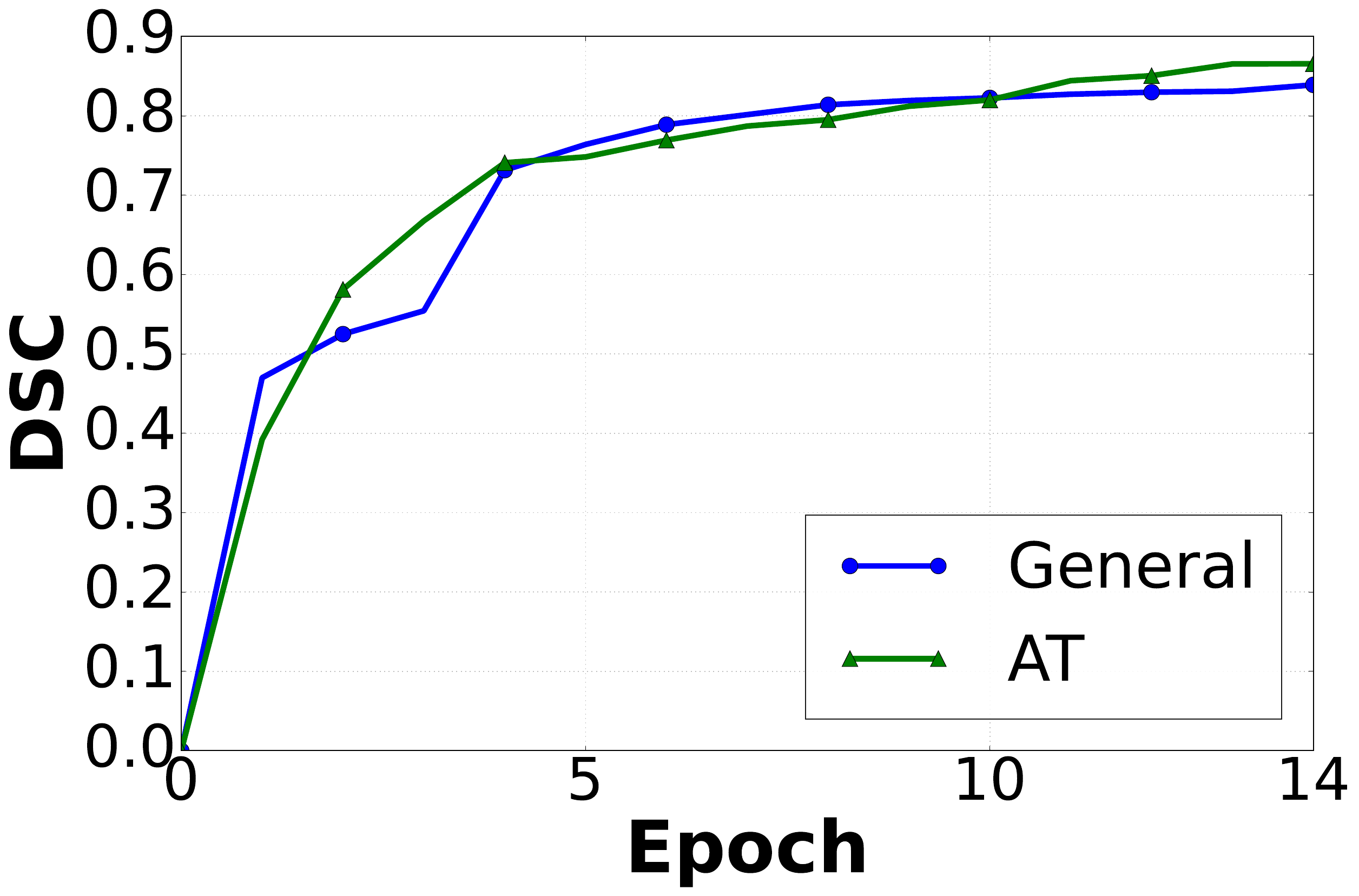}
    \subcaption{}\label{fig:b}
  \end{subfigure}

  \vspace{-4pt}

  % Row 2
  \begin{subfigure}[t]{0.488\linewidth}
    \centering
    \includegraphics[width=\linewidth]{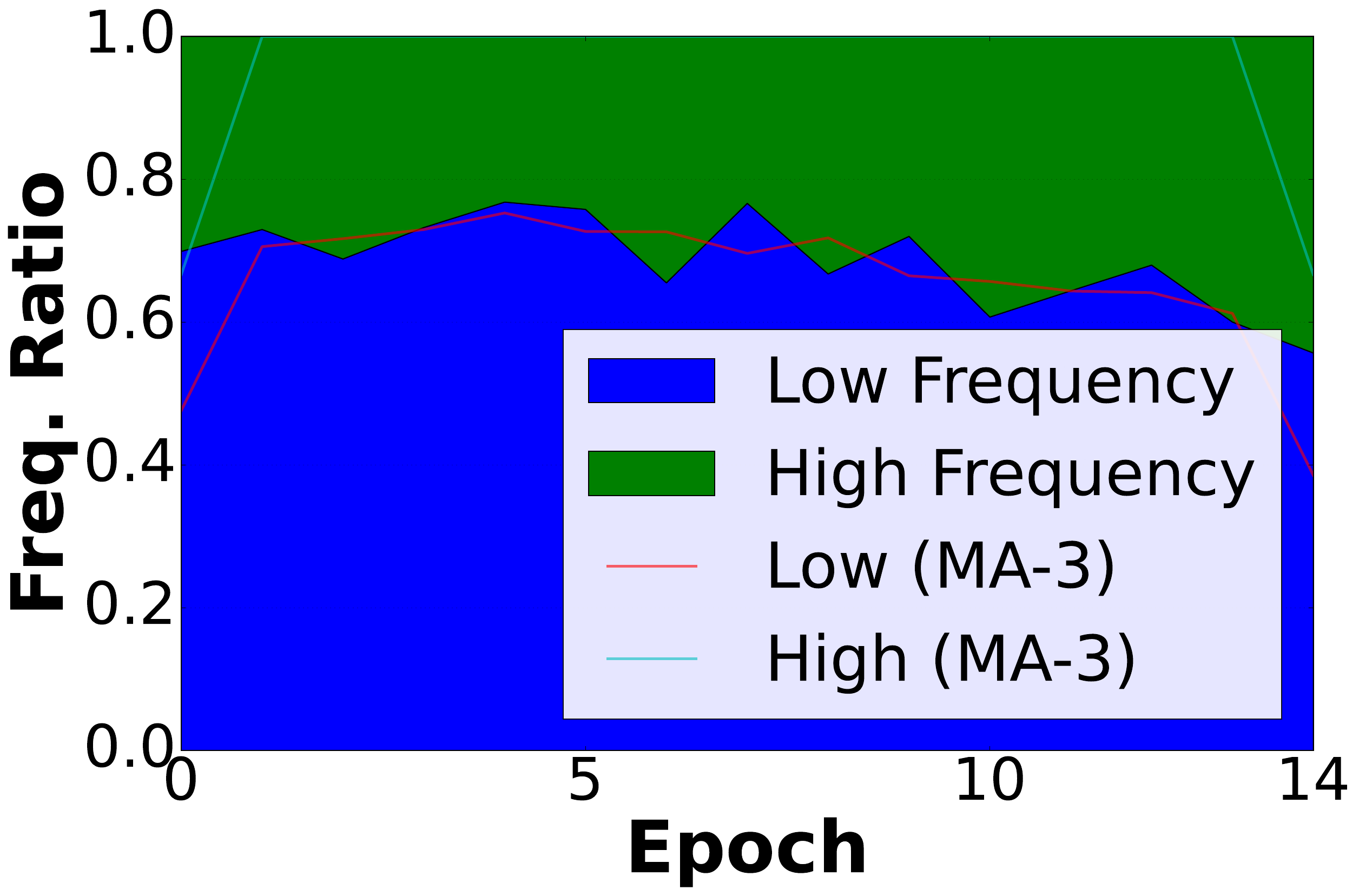}
    \subcaption{}\label{fig:c}
  \end{subfigure}\hspace{0.008\linewidth}
  \begin{subfigure}[t]{0.488\linewidth}
    \centering
    \includegraphics[width=\linewidth]{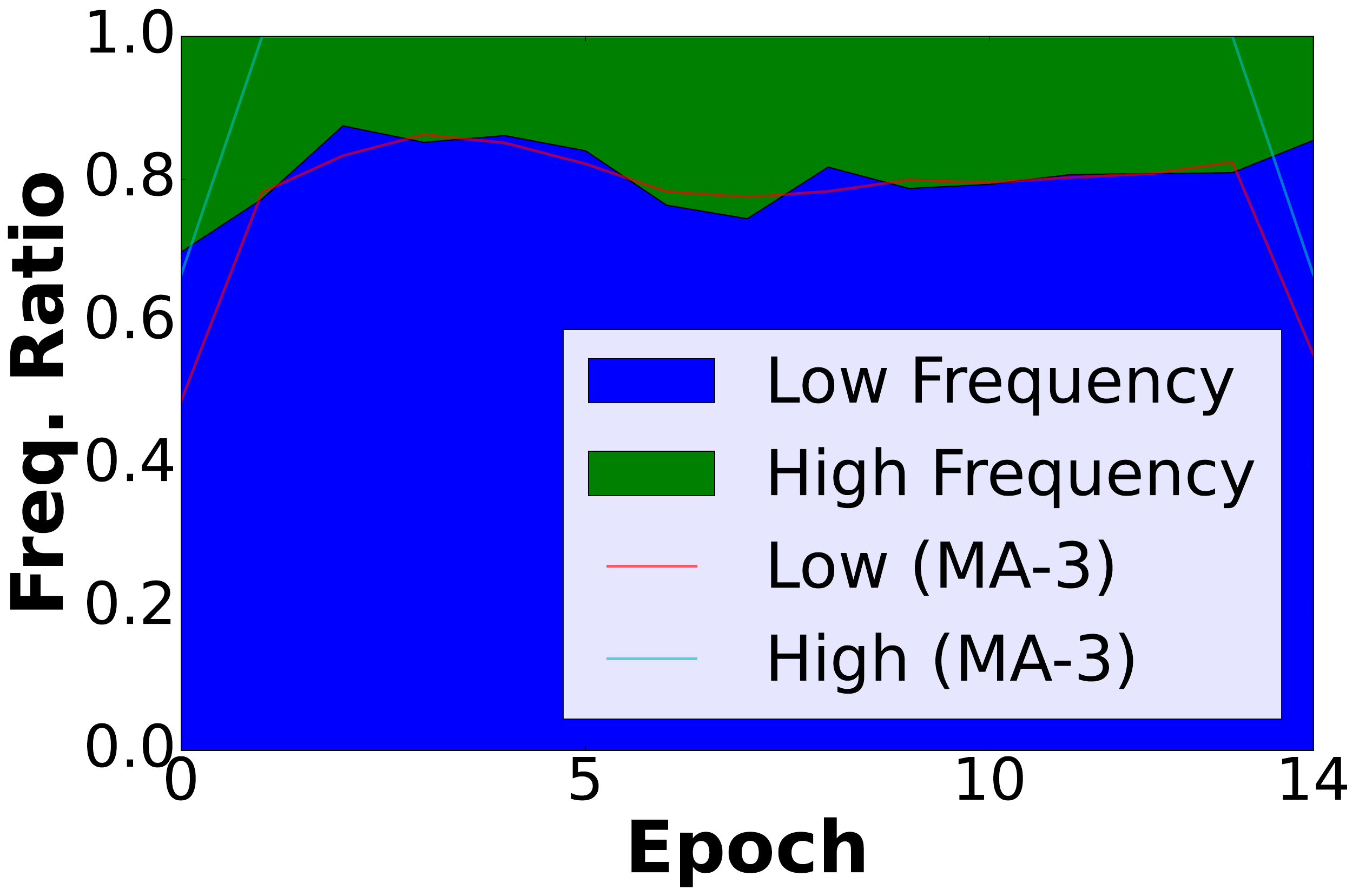}
    \subcaption{}\label{fig:d}
  \end{subfigure}

  \caption{Training dynamics and frequency allocation under clean/attacked inputs and general/adversarial training. (A) Dice with/without attack (shaded area: attack-induced drop). (B) Frequency ratio evolution in general training. (C–D) Frequency ratio in general vs. adversarial training. Colors and labels are aligned with the main text; green denotes adversarial/AT-related curves.}
  \label{fig:first}
\end{figure}

\IEEEPARstart{M}{edical} Medical image segmentation plays a pivotal role in clinical diagnostics, including tumor detection, disease diagnosis, and treatment planning \cite{8,9,10,17,zhao2024nfmpatt,xu2025scrnet,zhao2026pgpl}. 
Recent advancements in deep learning have significantly improved segmentation accuracy, but they exhibit vulnerability to adversarial attacks \cite{12,pang2024blinding}, as illustrated in Fig.~\ref{fig:first}(A). 
Studies have shown that even small perturbations in input data can lead to substantial errors in segmentation, this vulnerability poses potential risks of data manipulation, thereby threatening the integrity of medical data and the reliability of deep learning algorithms in critical medical applications.

\captionsetup[subfigure]{labelformat=simple,labelsep=space}
\renewcommand\thesubfigure{\Alph{subfigure}}

  % 若需要主 caption（你也可以保持无主 caption 以节省空间）
  % \caption{Performance comparison and frequency analysis of segmentation models under different conditions.}
  % \label{fig:first}
% \end{figure}

A mainstream route to robustness is adversarial training (AT), which mixes adversarial and clean samples during optimization \cite{PGD,FGSM}. 
While AT can harden models against attacks, it also alters the frequency allocation learned under standard training (Fig.~\ref{fig:first}B), driving a drift in spectral attention when the network is repeatedly exposed to adversarial perturbations (Fig.~\ref{fig:first}D). 
This drift stems from distribution and frequency mismatches between adversarial and natural inputs (highlighted in the late-stage view of Fig.~\ref{fig:first}C) and can be amplified by dataset-specific spectral characteristics \cite{16}.
Concretely, when attacks erode low-frequency content, the model is pushed toward high-frequency cues, weakening global structure; conversely, when attacks target high-frequency details, the model shifts to low-frequency components and neglects edges and fine boundaries. 
Regardless of direction, such training-induced spectral bias often degrades accuracy on clean images and adds nontrivial training and tuning overhead.

To address this challenge, we introduce \textbf{Layer-wise Noise-Guided Selective Wavelet Reconstruction (LNG-SWR)}, a dual-domain framework that couples training-time prior formation with input/feature-level execution. 
During training, we inject small zero-mean noise at multiple layers to learn a frequency-bias prior that steers representations away from noise-sensitive directions. 
Guided by this prior, we then perform selective wavelet reconstruction on the input/feature branch: preserve the low-frequency band, suppress the diagonal high-frequency band, and re-weight the horizontal/vertical bands before inverse reconstruction. 
This prior-guided gating delivers frequency adaptation—attenuating noise-sensitive components and enhancing directional structures and shape cues—thereby stabilizing boundary responses and maintaining spectral consistency. 
The design is backbone-agnostic and lightweight: layer-wise noise operates only at training time, and the wavelet module adds low inference overhead.
As a result, LNG-SWR can serve as a plug-in enhancement to adversarial training, while also improving robustness under standard (non-adversarial) training.

Comprehensive experiments conducted on three distinct medical image segmentation datasets provide strong evidence of our proposed model's superior performance compared to baseline models. Contributions of this paper are outlined below:

\begin{enumerate}
  \item We propose LNG-SWR, a network that follows a “train-time guidance, test-time execution” paradigm: small layer-wise noise shapes a prior during training, and selective wavelet reconstruction enforces it at inference. The design is plug-and-play, backbone-agnostic, and low-overhead, improving robustness without sacrificing clean accuracy.
  \item We introduce a layer-wise noise–guided selective wavelet reconstruction mechanism that performs band-wise preservation, suppression, and re-weighting in the wavelet domain to achieve more precise structure and boundary reconstruction. We also provide a systematic study that clarifies the roles of different frequency components in medical image segmentation.

  \item We establish a unified robustness protocol on CT (LIDC-IDRI) and ultrasound (TN-SCUI) with PGD-\(L_\infty/L_2\) and SSAH. LNG-SWR shows consistent clean Dice/IoU improvements and reduced performance drop under attacks, with additive gains when combined with AT on LIDC-IDRI.
\end{enumerate}

\section{Related Works}

\subsection{Medical Image Segmentation}
Deep convolutional architectures remain the mainstream for medical image segmentation. FCNs \cite{FCN} established end-to-end dense prediction, while U\textsc{-}Net \cite{unet} introduced an encoder–decoder with skip connections that recover spatial detail from coarse semantic features. Variants strengthen multi-scale representation and optimization under limited data: MultiResU\textsc{-}Net \cite{ibtehaz2020multiresunet} couples multi-resolution blocks with residual links; M$^2$SNet \cite{MSNet} replaces single-scale units with intra-layer multi-scale units to enlarge the receptive field; MShNet \cite{MShNet} integrates a pyramid-style fusion module to aggregate context at different scales. Recent CNN pipelines continue to report strong accuracy across modalities and tasks \cite{70,shim,18,wu2025multi}, benefiting from design choices such as dilated convolutions, depthwise separable kernels, attention gates, and lightweight decoders.

Despite progress, several characteristics of medical images remain challenging. Lesions can be small and have weak or fuzzy boundaries; class imbalance is severe; volumetric data are anisotropic and often low-contrast; and domain shifts arise from scanners, protocols, and institutions. To mitigate these factors, prior work explores boundary-aware losses, shape or topology constraints, multi-stage/cascade decoders, and 2D–3D hybrids that mix slice-wise efficiency with volumetric context. 
Frequency-aware components and explicit multi-scale fusion are also used to stabilize contours and suppress noise-sensitive responses. However, most designs are optimized for clean inputs and can be brittle under perturbations or adversarial noise, motivating architectures that control frequency responses and improve robustness with low overhead.

\subsection{Adversarial Attacks and Robustness}
Adversarial example generation has been studied extensively. Optimization-based methods include the box-constrained L\textsc{-}BFGS formulation \cite{3} and the C\&W attack \cite{4}. Gradient-based methods span FGSM \cite{FGSM}, its iterative variants BIM \cite{BIM} and PGD \cite{PGD}, and segmentation-oriented attacks such as SegPGD \cite{segPGD}. Frequency-aware attacks aim to control spectral properties of perturbations, e.g., spectrum simulation \cite{SSA} and high-frequency–focused strategies with low-frequency constraints \cite{SSAH}; universal frequency-tuned perturbations have also been explored \cite{87}. 

Adversarial training (AT)—mixing adversarial and clean samples during optimization—remains a strong empirical defense \cite{hardlm,FGSM,PGD}. Yet AT often introduces a clean–robustness trade-off, increases training and tuning cost, and is sensitive to attack and loss settings, especially in data- and compute-constrained medical scenarios. These limitations motivate defenses that improve robustness with lower overhead and better compatibility with standard training.

% \paragraph{Positioning of our approach.}
Unlike defenses that rely primarily on AT, our method couples layer-wise noise guidance with prior-guided selective wavelet reconstruction. Small zero-mean noise injected at multiple depths learns a frequency-bias prior that steers features away from noise-sensitive directions. Guided by this prior, selective wavelet gating preserves low-frequency content, suppresses diagonal high-frequency components, and re-weights directional bands at the input/feature level. This design forms the basis of our LNG\textnormal{-}SWR framework, which serves as a lightweight, backbone-agnostic plug-in that is compatible with both adversarial and standard training.

\section{Methods}
\begin{figure*}
\centerline{\includegraphics[width=0.968\textwidth]{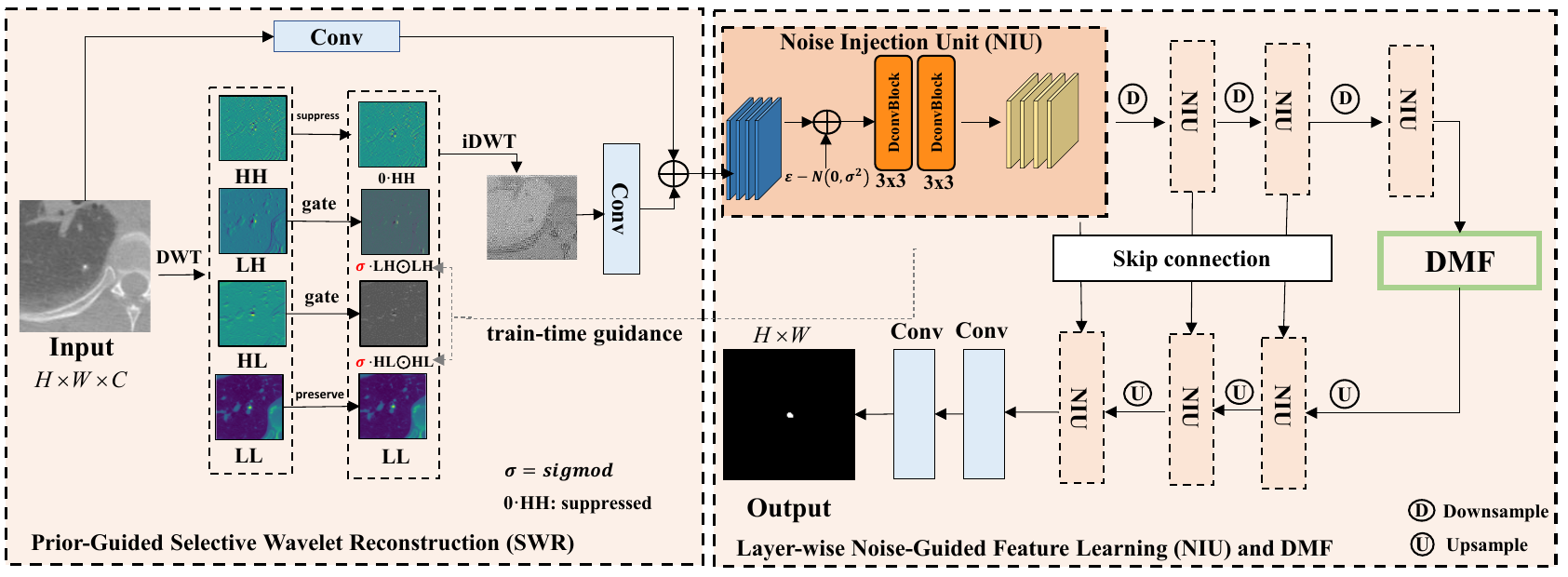}}
\caption{Overview of LNG-SWR. SWR (left): DWT splits the input into LL/LH/HL/HH; keep LL, suppress HH, and gate LH/HL with 1$\times$1\,+\,sigmoid ($\alpha$). 
iDWT reconstructs a branch that is 3$\times$3–refined and sum–fused ($\oplus$) with the raw branch, then sent to U\textsc{-}Net. 
NIU (train only) adds zero-mean noise inside conv blocks; it is off at test time. 
DMF (bottleneck) aggregates multi-scale context. 
\emph{Legend:} $\odot$ channel-wise product; $\oplus$ sum; $0{\times}\mathrm{HH}$\; suppressed;\quad
\raisebox{0.15ex}{\textcircled{\scriptsize D}}/\raisebox{0.15ex}{\textcircled{\scriptsize U}}\; down/upsample.}
\label{fig:fig2}
\end{figure*}

We present \emph{Layer-wise Noise-Guided Selective Wavelet Reconstruction (LNG-SWR)}. As shown in Figure~\ref{fig:fig2}, the method has three core designs: (i) \emph{layer-wise noise guidance} that injects small zero-mean noise at multiple depths layer to learn a frequency-bias prior; (ii) \emph{prior-guided selective wavelet reconstruction} on the input/feature branch that preserves low-frequency, suppresses diagonal high-frequency, and re-weights horizontal/vertical bands, followed by inverse reconstruction; (iii) a lightweight \emph{dynamic multi-scale fusion} (DMF) module at the bottleneck with strip convolutions for anisotropic context. The reconstructed branch is fused with the original branch (sum) and then processed by a standard U\textsc{-}Net encoder–decoder. The design is plug-in for U\textsc{-}Net–style backbones and adds low inference overhead.

We consider 2D medical images as tensors $x \in \mathbb{R}^{B\times C\times H\times W}$, where $B$ is batch size, $C$ the number of channels (e.g., $C{=}1$ for CT/ultrasound), and $H{\times}W$ the spatial size. Ground-truth labels are given either as dense integer maps $y \in \{0,\ldots,K\!-\!1\}^{B\times H\times W}$ or as one-hot tensors $Y \in \{0,1\}^{B\times K\times H\times W}$ for $K$ classes (binary segmentation is the special case $K{=}1$ with a sigmoid head). Unless otherwise stated, inputs are intensity-normalized per volume/slice.

\subsubsection{Prior-Guided Selective Wavelet Reconstruction (SWR)}
\label{subsec:swr}

% \noindent\textbf{English.}
We implement SWR as a light, differentiable sequence of four steps; shapes assume an input
$x\!\in\!\mathbb{R}^{B\times C\times H\times W}$ and orthonormal Haar wavelets.

We use a fixed, orthonormal 2D Haar DWT to split features into
$\{\mathrm{LL},\mathrm{LH},\mathrm{HL},\mathrm{HH}\}$.
It is implemented as depthwise $2{\times}2$ separable convolutions with stride~2,
and the inverse iDWT uses the transpose of the same filters;
the layer is differentiable with negligible FLOPs (reflection padding).

In 2D wavelet decomposition, the LL subband carries low-frequency structure and contrast, 
whereas LH/HL encode vertical/horizontal edges, and HH aggregates diagonal fine detail~\cite{sutha2013comprehensive}.
Medical images---especially ultrasound---exhibit noise with strong high-frequency characteristics (e.g., speckle), 
which tends to populate the detail subbands, with diagonal HH being particularly noise-sensitive; 
wavelet-domain denoising commonly thresholds or attenuates such components~\cite{chen2006wavelet}. 
Meanwhile, adversarial perturbations often manifest as high-frequency energy in the spectrum, 
degrading recognition while minimally affecting perceptual low-frequency content~\cite{yin2019fourier}. 
Therefore, we preserve LL to retain semantic layout, re-weight LH/HL to keep 
directional edges under learnable control, and suppress HH to curb noise-prone diagonal detail 
before iDWT reconstruction. This design stabilizes boundary responses while avoiding global over-smoothing.

\paragraph{2D DWT (depthwise, stride-2).}
Let $h=[1,1]/\sqrt{2}$ and $g=[1,-1]/\sqrt{2}$.
Define four $2{\times}2$ separable kernels
$K_{\text{LL}}=h^\top h,\; K_{\text{LH}}=h^\top g,\; K_{\text{HL}}=g^\top h,\; K_{\text{HH}}=g^\top g$.
We apply depthwise convolutions with stride $2$:
\begin{align}
\mathbf{X}_{\text{LL}} &= \mathrm{DWConv}\!\left(x, K_{\text{LL}}, s{=}2\right),\\
\mathbf{X}_{\text{LH}} &= \mathrm{DWConv}\!\left(x, K_{\text{LH}}, s{=}2\right),\\
\mathbf{X}_{\text{HL}} &= \mathrm{DWConv}\!\left(x, K_{\text{HL}}, s{=}2\right),\\
\mathbf{X}_{\text{HH}} &= \mathrm{DWConv}\!\left(x, K_{\text{HH}}, s{=}2\right).
\end{align}

giving sub-bands $\mathbf{X}_{\bullet}\!\in\!\mathbb{R}^{B\times C\times \frac{H}{2}\times \frac{W}{2}}$.

\paragraph{Band-wise gating (learnable, per-channel).}
We preserve $\mathbf{X}_{\text{LL}}$ and suppress $\mathbf{X}_{\text{HH}}$.
For directional bands, we use $1{\times}1$ convs followed by a sigmoid gate:
\begin{align}
\widehat{\mathbf{X}}_{\text{LH}}
=\sigma(\mathbf{W}_{\text{LH}}\!\ast_{1\times1}\!\mathbf{X}_{\text{LH}})\odot \mathbf{X}_{\text{LH}}, \\
\quad
\widehat{\mathbf{X}}_{\text{HL}}
=\sigma(\mathbf{W}_{\text{HL}}\!\ast_{1\times1}\!\mathbf{X}_{\text{HL}})\odot \mathbf{X}_{\text{HL}},
\end{align}
where $\mathbf{W}_{\text{LH}},\mathbf{W}_{\text{HL}}\!\in\!\mathbb{R}^{C\times C\times 1\times 1}$ (often diagonal to stay depthwise), $\sigma$ is the sigmoid, and $\odot$ is Hadamard product. We set $\widehat{\mathbf{X}}_{\text{HH}}=\mathbf{0}$ and $\widehat{\mathbf{X}}_{\text{LL}}=\mathbf{X}_{\text{LL}}$.

\paragraph{iDWT (depthwise transposed conv, stride-2).}
Using the same Haar kernels (orthonormal case), reconstruct:
\begin{align}
\tilde{x}
=\mathrm{DWConv}^{\!\top}(\widehat{\mathbf{X}}_{\text{LL}},K_{\text{LL}},s{=}2) \\
+\mathrm{DWConv}^{\!\top}(\widehat{\mathbf{X}}_{\text{LH}},K_{\text{LH}},s{=}2)\\
+\mathrm{DWConv}^{\!\top}(\widehat{\mathbf{X}}_{\text{HL}},K_{\text{HL}},s{=}2)\\
+\mathrm{DWConv}^{\!\top}(\widehat{\mathbf{X}}_{\text{HH}},K_{\text{HH}},s{=}2),
\end{align}
yielding $\tilde{x}\!\in\!\mathbb{R}^{B\times C\times H\times W}$.

\paragraph{Shallow conv \& fusion.}
A $3{\times}3$ conv–BN–ReLU refines the reconstructed signal:
\begin{align}
\tilde{f}_0=\mathrm{ReLU}\!\big(\mathrm{BN}(\mathrm{Conv}_{3\times3}(\tilde{x}))\big), \\
f_0=\mathrm{ReLU}\!\big(\mathrm{BN}(\mathrm{Conv}_{3\times3}(x))\big).
\end{align}
We use \emph{sum fusion} before entering the encoder: $f_0^{\oplus}=f_0+\tilde{f}_0$ (same channel width).

\noindent\emph{Kernel sizes and cost.}
DWT/iDWT use fixed $2{\times}2$ depthwise (transposed) convolutions with stride $2$;
gating uses $1{\times}1$; the refinement conv is $3{\times}3$.
All ops are GPU-friendly and add $<\!5\%$ FLOPs in our settings.

\subsubsection{Noise Injection Unit (NIU)}
\label{subsec:niu}

% \noindent\textbf{English.}
As shown in Algorithm 2, NIU injects small zero-mean Gaussian noise into feature maps \emph{inside convolutional blocks} during training, forming a frequency-bias prior while adding negligible inference cost (disabled at test time). 
For the $l$-th stage with feature
$\mathbf{h}^{(l)}\!\in\!\mathbb{R}^{B\times C_l\times H_l\times W_l}$, NIU operates as:
\[
\begin{aligned}
&\text{Noise sampling:}\quad &&\boldsymbol{\epsilon}^{(l)} \sim \mathcal{N}\!\left(\mathbf{0},\, (\sigma_l^2)\,\mathbf{I}\right),\\
&\text{Feature perturbation:}\quad &&\widehat{\mathbf{h}}^{(l)} \;=\; \mathbf{h}^{(l)} + \boldsymbol{\epsilon}^{(l)},\\
&\text{Block update:}\quad &&\mathbf{h}^{(l+1)} \;=\; f^{(l)}\!\big(\widehat{\mathbf{h}}^{(l)}\big),
\end{aligned}
\]
where $f^{(l)}$ denotes a \emph{DoubleConv} block ($3{\times}3$–BN–ReLU, $3{\times}3$–BN–ReLU). 
To keep the perturbation small and stage-aware, we use a per-stage magnitude
\[
\begin{aligned}
&\sigma_l \;=\; \gamma_l\,\sigma_0,\qquad \gamma_l\in(0,1],\ \sigma_0>0,
\end{aligned}
\]
and optionally a channel-wise gate (diagonal $1{\times}1$) to modulate noise strength:
\[
\begin{aligned}
&\boldsymbol{\epsilon}^{(l)}_{\!c} \;=\; \sigma\!\big(\alpha^{(l)}_{c}\big)\,\boldsymbol{\eta}^{(l)}_{\!c},\qquad 
\boldsymbol{\eta}^{(l)}_{\!c}\sim \mathcal{N}(0,1),\ \ c=1,\ldots,C_l,
\end{aligned}
\]
where $\sigma(\cdot)$ is sigmoid and $\alpha^{(l)}_{c}$ are learnable scalars (default: fixed $\sigma_l$ when gates are not used). 
During inference we set $\sigma_l\!=\!0$ for all $l$:
\[
\begin{aligned}
\text{eval mode:}\qquad \widehat{\mathbf{h}}^{(l)} \;=\; \mathbf{h}^{(l)},\qquad \mathbf{h}^{(l+1)} \;=\; f^{(l)}\!\big(\mathbf{h}^{(l)}\big).
\end{aligned}
\]
NIU is applied at shallow encoders and each encoder/decoder stage; it is \emph{not} used inside the SWR block. 
This training-time perturbation nudges representations away from noise-sensitive directions, and the induced prior is executed by the SWR band gating.

\subsubsection{Encoder / Downsampling Stages}
\label{subsec:down}

% \noindent\textbf{English.}
Starting from the fused shallow feature $f_0^{\oplus}\!\in\!\mathbb{R}^{B\times C_0\times H\times W}$, the encoder stacks $S$ downsampling stages. 
Each stage halves spatial resolution and (optionally) doubles channel width, and applies NIU (train only) inside the convolutional block.
Let $(H_s,W_s)=(H/2^{s},\,W/2^{s})$ and $C_s=\kappa_s C_0$ (typically $\kappa_s=2^{s}$ for U\textsc{-}Net). 
For stage $s=1,\ldots,S$:

\begin{IEEEeqnarray}{rCl}
\mathbf{z}_s & = & \mathrm{MaxPool}_{2\times2}\!\big(\mathbf{h}_{s-1}\big), \qquad \mathbf{h}_0=f_0^{\oplus} \IEEEyesnumber\IEEEyessubnumber\\
\widehat{\mathbf{z}}_s & = & \mathbf{z}_s + \boldsymbol{\epsilon}_s,\quad 
\boldsymbol{\epsilon}_s \sim \mathcal{N}\!\big(\mathbf{0},\,\sigma_s^2\mathbf{I}\big) \IEEEyessubnumber\\
\mathbf{h}_s & = & \phi_{\text{enc}}^{(s)}\!\big(\widehat{\mathbf{z}}_s\big) \IEEEyessubnumber\\
\mathbf{h}_s & \in & \mathbb{R}^{B\times C_s\times H_s\times W_s},\quad 
(H_s,W_s)=\!\left(\tfrac{H}{2^s},\,\tfrac{W}{2^s}\right) \IEEEyessubnumber
\end{IEEEeqnarray}

We store skip features $\{\mathbf{h}_1,\ldots,\mathbf{h}_{S-1}\}$ for the decoder. 
During inference, NIU is disabled ($\widehat{\mathbf{z}}_s=\mathbf{z}_s$). 
The design matches standard U\textsc{-}Net encoders while integrating layer-wise noise guidance without altering tensor interfaces, keeping FLOPs overhead small.

\begin{figure}
\centerline{\includegraphics[width=0.96\columnwidth]{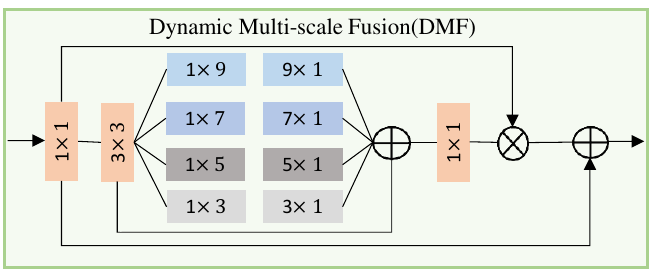}}
\caption{Dynamic Multi-Scale Fusion (DMF). Parallel strip convolutions capture anisotropic context. 
Default branches use $1{\times}3,\,1{\times}5,\,3{\times}1,\,5{\times}1$; an optional long-strip variant extends to $1{\times}7,\,1{\times}9,\,7{\times}1,\,9{\times}1$. 
Branch outputs are concatenated, projected by $1{\times}1$, optionally SE-gated, and added residually. 
Spatial size is preserved with small computational cost.}

\label{fig:fig3}
\end{figure}

\subsubsection{Dynamic Multi-Scale Fusion (DMF)}
\label{subsec:dmf}
DMF is a lightweight bottleneck that aggregates \emph{anisotropic} context with horizontal/vertical strip convolutions while keeping spatial size unchanged. 
Given a bottleneck feature $\mathbf{h}_B\!\in\!\mathbb{R}^{B\times C_B\times H_B\times W_B}$, we apply a small set of parallel strip branches and fuse them with a $1{\times}1$ projection and an optional residual.

\noindent\textbf{Branches (merged form).}
Let the kernel set be

\begin{equation} 
\begin{aligned}
\mathcal{K} &= \{\,1{\times}3,\;1{\times}5,\;3{\times}1,\;5{\times}1\,\} \\ 
&\quad \text{optional long strips: } (1{\times}7,\,1{\times}9,\,7{\times}1,\,9{\times}1).
\end{aligned}
\end{equation}

% \end{aligned}

For each $k\!\in\!\mathcal{K}$, one branch is a Conv–BN–ReLU with kernel size $k$ and intermediate width $C_I$.
Collecting all branches:
\[
\mathbf{u} \;=\; \big\|_{k\in\mathcal{K}} \mathrm{ReLU}\!\big(\mathrm{BN}(\mathrm{Conv}_k(\mathbf{h}_B))\big)
\ \in\ \mathbb{R}^{B\times (|\mathcal{K}|\,C_I)\times H_B\times W_B}.
\]
A $1{\times}1$ projection restores the channel width and (optionally) adds a residual:
\begin{equation} 
\mathbf{y}=\mathrm{ReLU}\!\big(\mathrm{BN}(\mathrm{Conv}_{1\times1}(\mathbf{u}))\big),\qquad
\mathbf{f}_B=\mathbf{y}+\mathbf{h}_B.
\end{equation}

\noindent\textbf{Kernel and width settings.}
We use $\mathcal{K}=\{1{\times}3,1{\times}5,3{\times}1,5{\times}1\}$ by default; when resolution permits, we extend to long strips $\{1{\times}7,1{\times}9,7{\times}1,9{\times}1\}$ to further enlarge directional receptive fields with modest FLOPs growth. 
To keep overall parameters stable, set $C_I=C_B/|\mathcal{K}|$. 
DMF complements SWR: the former supplies orientation-aware spatial context, the latter stabilizes spectrum usage.

\subsubsection{Decoder / Upsampling Stages}
\label{subsec:up}

% \noindent\textbf{English.}
Starting from the bottleneck feature $\mathbf{f}_B\!\in\!\mathbb{R}^{B\times C_S\times H_S\times W_S}$ (after DMF), the decoder mirrors the encoder with $S$ upsampling stages. Each stage upsamples by a factor of $2$, concatenates the corresponding skip feature, applies NIU (train only), and refines with a DoubleConv block. Let skips be $\{\mathbf{h}_{S-1},\ldots,\mathbf{h}_1\}$ from the encoder and define output spatial sizes $(\tilde H_t,\tilde W_t)=(H/2^{t},\,W/2^{t})$. For $t=S-1,\ldots,1$:
\begin{IEEEeqnarray}{rCl}
\mathbf{u}_{t+1} &\in& \mathbb{R}^{B\times C_{t+1}\times \tilde H_{t+1}\times \tilde W_{t+1}},\qquad \mathbf{u}_{S}=\mathbf{f}_B \IEEEyesnumber\IEEEyessubnumber\\
\mathbf{u}_{t}^{\uparrow} &=& 
\begin{cases}
\mathrm{Upsample}_{\text{bilinear}\times 2}(\mathbf{u}_{t+1}), & \text{(default)}\\
\mathrm{ConvT}_{2\times2,\,s=2}(\mathbf{u}_{t+1}), & \text{(optional)}
\end{cases} \IEEEyessubnumber\\
\mathbf{z}_{t} &=& [\,\mathbf{h}_{t}\ \|\ \mathbf{u}_{t}^{\uparrow}\,] \;\in\; \mathbb{R}^{B\times (C_t+C_{t+1}/2)\times \tilde H_{t}\times \tilde W_{t}} \IEEEyessubnumber\\
\widehat{\mathbf{z}}_{t} &=& \mathbf{z}_{t} + \boldsymbol{\epsilon}_{t},\quad 
\boldsymbol{\epsilon}_{t}\sim \mathcal{N}\!\big(\mathbf{0},\,\sigma^{2}_{t}\mathbf{I}\big) \quad \text{(NIU, train only)} \IEEEyessubnumber\\
\mathbf{u}_{t} &=& \phi^{(t)}_{\text{dec}}(\widehat{\mathbf{z}}_{t}) 
\;=\; \mathrm{DConv}_{3\times3}(\widehat{\mathbf{z}}_{t}) 
\ \in\ \mathbb{R}^{B\times C_{t}\times \tilde H_{t}\times \tilde W_{t}}. \IEEEyessubnumber
\end{IEEEeqnarray}
Finally, the prediction head applies a $1{\times}1$ projection:
\begin{align*}
\mathbf{p}\;=\;\mathrm{Conv}_{1\times1}(\mathbf{u}_{1})\in\mathbb{R}^{B\times K\times H\times W},\\
\hat{\mathbf{y}}=\mathrm{softmax/sigmoid}(\mathbf{p}).
\end{align*}
By injecting NIU only inside convolutional blocks and using bilinear upsampling by default, the decoder keeps inference overhead low while leveraging skip connections for detail recovery.
% ---------- Algorithm 1 (algorithmic 版本) ----------
\begin{algorithm}[t]
\caption{Layer-wise Noise-Guided Training (NIU)}
\label{alg:niu}
\begin{algorithmic}[1]
\REQUIRE Training set $\mathcal{D}{=}\{(x,y)\}$; network $f_\theta$ with features $\{F^{(l)}\}_{l=1}^L$; noise scale $\sigma$; loss $\mathcal{L}_{\mathrm{seg}}$; optimizer $\mathrm{Opt}$; epochs $E$; batch size $B$; (optional) AT steps $K$, step size $\alpha$, budget $\epsilon$.
\ENSURE Parameters $\theta^\star$ with robustness-oriented frequency bias.
\FOR{$e{=}1$ \TO $E$}
  \FOR{each mini-batch $\{(x_i,y_i)\}_{i=1}^B \subset \mathcal{D}$}
    \STATE \textbf{(Optional AT)} $x^{\mathrm{adv}}{\leftarrow}x$
    \IF{use-AT}
      \FOR{$k{=}1$ \TO $K$}
        \STATE $g \leftarrow \nabla_{x}\,\mathcal{L}_{\mathrm{seg}}\!\big(f_\theta(x^{\mathrm{adv}}),y\big)$
        \STATE $x^{\mathrm{adv}} \leftarrow \Proj_{\mathcal{B}(x,\epsilon)}\!\big(x^{\mathrm{adv}} + \alpha\,\mathrm{sign}(g)\big)$
      \ENDFOR
    \ENDIF
    \IF{use-AT}
      \STATE $x' \leftarrow x^{\mathrm{adv}}$
    \ELSE
      \STATE $x' \leftarrow x$
    \ENDIF
    \FOR{$l{=}1$ \TO $L$}
      \STATE Compute $F^{(l)}$; sample $\varepsilon^{(l)} \!\sim\! \mathcal{N}(0,\sigma^2 I)$
      \STATE $\tilde{F}^{(l)} \leftarrow F^{(l)} + \varepsilon^{(l)}$ \COMMENT{train-time only}
      \STATE Pass $\tilde{F}^{(l)}$ forward
    \ENDFOR
    \STATE $\hat{y} \leftarrow f_\theta(x')$;\quad $\mathcal{L} \leftarrow \mathcal{L}_{\mathrm{seg}}(\hat{y},y)$
    \STATE (optional) $\mathcal{L} \mathrel{+}= \lambda \sum_{l}\!\big\|\phi(F^{(l)})-\phi(\tilde{F}^{(l)})\big\|_2^2$
    \STATE $\theta \leftarrow \mathrm{Opt}\!\big(\theta, \nabla_{\theta}\mathcal{L}\big)$
  \ENDFOR
\ENDFOR
\STATE \textbf{return} $\theta^\star$
\end{algorithmic}
\end{algorithm}

% ---------- Algorithm 2 (algorithmic 版本) ----------
\begin{algorithm}[t]
\caption{Band-Selective Wavelet Reconstruction (SWR, inference)}
\label{alg:swr}
\begin{algorithmic}[1]
\REQUIRE Feature map $Z$; 2D Haar $\DWT/\iDWT$; gates $G_{ll},G_{lh},G_{hl},G_{hh}\!\in[0,1]$ (or predictor $g$).
\ENSURE Reconstructed feature $\hat{Z}$.
\STATE $(X_{ll},X_{lh},X_{hl},X_{hh}) \leftarrow \DWT(Z)$
\IF{static}
  \STATE $G_{ll}{\leftarrow}1$;\ $G_{lh},G_{hl}{\leftarrow}\sigm(\alpha_{lh}),\sigm(\alpha_{hl})$;\ $G_{hh}{\leftarrow}0$
\ELSE
  \STATE $(G_{lh},G_{hl},G_{hh}) \leftarrow \sigm\big(g(X_{ll})\big)$;\ $G_{ll}{\leftarrow}1$
\ENDIF
\STATE $\hat{Z} \leftarrow \iDWT\!\big(X_{ll},\, G_{lh}{\odot}X_{lh},\, G_{hl}{\odot}X_{hl},\, G_{hh}{\odot}X_{hh}\big)$
\STATE \textbf{return} $\hat{Z}$
\end{algorithmic}
\end{algorithm}

\subsubsection{Training Objective (Loss)}
\label{subsec:loss}

% \noindent\textbf{English.}
We follow the paper and use the standard Dice + Cross-Entropy objective without any additional regularizers. 
Given logits $\mathbf{p}\in\mathbb{R}^{B\times K\times H\times W}$ and predictions 
$\hat{\mathbf{y}}=\mathrm{softmax}(\mathbf{p})$ for $K{>}1$ (or $\hat{\mathbf{y}}=\sigma(\mathbf{p})$ for $K{=}1$), 
with one-hot labels $Y\in\{0,1\}^{B\times K\times H\times W}$ and a small $\varepsilon>0$, we define:
\begin{align}
\mathcal{L}_{\text{CE}} 
&= -\frac{1}{BHW}\sum_{b=1}^{B}\sum_{i=1}^{H}\sum_{j=1}^{W}\sum_{k=1}^{K}
Y_{b,k,i,j}\,\log\big(\hat{y}_{b,k,i,j}\big), \label{eq:ce_main}\\
\mathcal{L}_{\text{Dice}}
&= 1 - \frac{1}{K}\sum_{k=1}^{K}
\frac{2\sum_{b,i,j}\hat{y}_{b,k,i,j}Y_{b,k,i,j}+\varepsilon}
{\sum_{b,i,j}\hat{y}_{b,k,i,j}+\sum_{b,i,j}Y_{b,k,i,j}+\varepsilon}. \label{eq:dice_main}
\end{align}
The total loss is:
\begin{align}
\mathcal{L} \;=\; \mathcal{L}_{\text{CE}} \;+\; \lambda_{\text{dice}}\,\mathcal{L}_{\text{Dice}}.
\end{align}
For binary segmentation we use the sigmoid version of \eqref{eq:ce_main}–\eqref{eq:dice_main} with $K{=}1$.

\section{Experiments and Results}
\subsection{Experimental Setup}
We implemented all algorithms using the PyTorch framework and conducted training on NVIDIA Tesla V100 32GB GPUs. For optimization during training, we chose Root Mean Square Propagation (RMSProp), with a batch size of 4, a weight decay of $1e-8$, and a momentum of 0.9. The initial learning rate was set at $1e-4$, and all models were trained from scratch, without any pre-training. For the lung nodule and liver segmentation dataset $\sigma=0.03$, while for the thyroid nodule dataset, $\sigma$ is set as $0.1$.
To optimize computational resources, we adopted a train-validate-test pattern for other datasets. 
% In terms of evaluation metrics, we primarily used the Dice Similarity Coefficient (DSC) and Intersection over Union (IoU). Additionally, to assess the models' performance in recognizing positive samples, we employed sensitivity (Sen) and Positive Predictive Value (PPV).
\subsection{Evaluation Metrics}
Let $P$ and $G$ denote the predicted and ground-truth binary masks, and let $TP,FP,FN$ be the pixel counts of true positives, false positives, and false negatives.

\paragraph{Dice similarity coefficient (DSC)}
\[
\mathrm{DSC} \;=\; \frac{2\,|P\cap G|}{|P|+|G|} \;=\; \frac{2\,TP}{2\,TP+FP+FN}.
\]
Dice measures overlap between two sets and is widely used in medical image segmentation; it is related to Jaccard/IoU via $\mathrm{DSC}=\tfrac{2\,\mathrm{IoU}}{1+\mathrm{IoU}}$. 

\paragraph{Intersection over Union (IoU)}
\[
\mathrm{IoU} \;=\; \frac{|P\cap G|}{|P\cup G|} \;=\; \frac{TP}{TP+FP+FN}.
\]
IoU is the Jaccard index, i.e., intersection divided by union. 

\paragraph{Sensitivity (Sen) and Positive Predictive Value (PPV)}
\[
\mathrm{Sen}\;{=}\;\frac{TP}{TP+FN}, \qquad
\mathrm{PPV}\;{=}\;\frac{TP}{TP+FP}.
\]
Sensitivity (recall, true positive rate) quantifies lesion detection, whereas PPV (precision) reflects false-positive control. 
\paragraph{Reporting protocol}
Unless otherwise specified, we compute per-image DSC/IoU/Sen/PPV on 2D slices and report the dataset mean (and standard deviation where noted). Thresholding and post-processing follow the same settings across methods to ensure comparability.

\subsection{Datasets}

% \noindent\textbf{LIDC–IDRI (Lung nodules).}
% We evaluate on the LIDC–IDRI repository. After filtering, 986 nodules annotated by four radiologists are retained. 
% To address inter–observer variability, we adopt the 50\% consensus rule~\cite{21} to derive the ground‐truth boundary. 
% Each nodule patch is cropped and resized to $128\times128$. We randomly split the nodules into 788 for training and 194 for testing.

% \medskip
% \noindent\textbf{Liver tumor MRI.}
% MRI scans from 327 liver–tumor patients are used (IRB approved; anonymized). 
% All images are resized to $320\times224$. Tumors are randomly split into 195/66/66 for training/validation/testing.

% \medskip
% \noindent\textbf{Thyroid nodules (TN–SCUI 2020).}
% We use the MICCAI 2020 TN–SCUI ultrasound set with 3{,}644 images (collected in the United States). 
% Images are cropped using label center points and resized to $256\times256$. 
% We adopt a random split of 60\%/20\%/20\% for training/validation/testing.
\noindent\textbf{LIDC--IDRI (lung nodules).}
We use the public LIDC--IDRI repository. After excluding scans with missing masks or ambiguous consensus, 986 nodules annotated by four radiologists are retained~\cite{armato2011lung}. 
To mitigate inter–observer variability, we follow the 50\% consensus protocol~\cite{21}: a pixel is positive if at least two radiologists mark it as lesion; the final mask is the largest connected component inside the consensus contour. 
Each nodule is cropped by a tight bounding box with a 10\,px margin and resized to $128{\times}128$ using bilinear (image) and nearest-neighbor (mask) interpolation. 
For intensity normalization, CT voxels are clipped to $[-1000,400]$ HU and min–max normalized to $[0,1]$ per volume. 
We perform a \emph{patient-level} split to avoid leakage: 788 nodules for training and 194 for testing; the test set contains only patients unseen during training. 
Online data augmentation includes random horizontal flip ($p{=}0.5$), $\pm10^\circ$ rotation, and scale jitter in $[0.9,1.1]$.

\medskip
\noindent\textbf{Liver tumor MRI.}
An in-house liver MRI cohort of 327 patients was curated under institutional IRB approval; all data were de-identified prior to analysis.
Studies with corrupted slices or incomplete annotations were excluded.
Axial slices were resampled to 1.0\,mm in-plane resolution (bicubic for images, nearest-neighbor for masks), z-score normalized per volume, and resized to $320{\times}224$.
A patient-level split of 195/66/66 patients was used for training/validation/testing to preclude cross-patient leakage.
Data augmentation followed the CT configuration with additional intensity jitter ($\pm 5\%$) to accommodate MRI contrast variability.
The dataset is not publicly available due to privacy restrictions; access can be arranged upon reasonable request and institutional approval.

\medskip
\noindent\textbf{Thyroid nodules (TN--SCUI).}
We use the MICCAI 2020 TN--SCUI ultrasound set (3{,}644 images, US centers). 
Images are cropped around label centroids with a fixed context window and resized to $256{\times}256$ (bilinear/nearest)~\cite{TNSCUI2020_Proceedings}. 
Speckle-preserving normalization is applied by percentile clipping (1st--99th) and per-image min–max scaling. 
A random 60\%/20\%/20\% split is used for training/validation/testing; when multiple images originate from the same study, they are kept within the same split. 
Augmentation includes flip, small rotation ($\pm8^\circ$) and elastic deformation ($\alpha{=}10$, $\sigma{=}3$) to mimic probe-induced distortions.

\subsection{Baselines}
We compare LNG\textendash SWR with eight representative segmentation networks:

\begin{enumerate}
  \item \textbf{U\mbox{-}Net}~\cite{unet}: symmetric encoder–decoder with skip connections for context capture and precise localization.
  \item \textbf{Attention U\mbox{-}Net}~\cite{AttUNet}: inserts attention gates on skip paths to suppress irrelevant responses and highlight targets.
  \item \textbf{DS\mbox{-}TransUNet}~\cite{DSTransUNet}: U\mbox{-}shaped model with hierarchical Swin Transformers in encoder/decoder to model global context.
  \item \textbf{TransAttUNet}~\cite{transattunet}: combines Transformer self/global attention with multi\mbox{-}level guided attention in a U\mbox{-}Net.
  \item \textbf{MT\mbox{-}UNet}~\cite{MTUNet}: mixed\mbox{-}Transformer U\mbox{-}Net using local–global attention for long\mbox{-}range dependency modeling.
  \item \textbf{UNeXt}~\cite{UNeXt}: lightweight Conv–MLP architecture aimed at rapid, low\mbox{-}compute medical segmentation.
  \item \textbf{CMUNeXt}~\cite{CMUNet}: efficient fully\mbox{-}convolutional network leveraging large kernels and inverted bottlenecks.
  \item FreqUNet~\cite{li2025frequnet} : U\mbox{-}shaped model with Visual State Space (Mamba) blocks for efficient long\mbox{-}range context modeling.

\item \textbf{FreqUNet}~\cite{li2025frequnet}: lightweight dual\mbox{-}branch with DWT split and S6/Fourier cues.
\item \textbf{UKAN}~\cite{li2025u}: U\mbox{-}Net backbone with KAN layers for non\mbox{-}linear token modeling.

\end{enumerate}

\begin{table*}[t]
\centering
\caption{Unified clean-set comparison across three datasets (TN--SCUI, LIDC-IDRI, Liver). 
All values are \%. Best per column in \textbf{bold}.}
\label{tab:exp1_unified_clean}
\small
\setlength{\tabcolsep}{10pt}
\renewcommand{\arraystretch}{1.15}
\begin{tabular}{l|cccc|cccc|cccc}
\hline
& \multicolumn{4}{c|}{\textbf{TN--SCUI}} 
& \multicolumn{4}{c|}{\textbf{LIDC-IDRI}} 
& \multicolumn{4}{c}{\textbf{Liver}}\\
\cline{2-13}
\textbf{Method} 
& DSC & IoU & Sen & PPV 
& DSC & IoU & Sen & PPV 
& DSC & IoU & Sen & PPV \\
\hline
UNet~\cite{unet} 
& 83.0 & 73.8 & 84.0 & 88.3
& 86.7 & 77.5 & 87.6 & 88.6
& 76.4 & 68.6 & 81.2 & 77.9 \\
AttUNet~\cite{AttUNet} 
& 85.2 & 75.9 & 87.1 & 87.3
& 86.9 & 77.7 & 88.3 & 88.0
& 76.7 & 69.0 & 78.5 & 80.8 \\
DeepLabV3~\cite{DeeplabV3} 
& 88.2 & 80.5 & 87.2 & 92.5
& 85.3 & 75.9 & 86.6 & 87.7
& 76.8 & 69.0 & \textbf{82.6} & 75.9 \\
CPFNet~\cite{CPFNet} 
& 88.7 & 81.3 & 89.9 & 90.6
& 85.5 & 75.8 & 86.8 & 87.4
& 74.1 & 66.2 & 81.0 & 75.1 \\
UNeXt~\cite{UNeXt} 
& 86.5 & 78.2 & 88.5 & 88.7
& 84.8 & 74.8 & 86.4 & 86.7
& 65.9 & 56.9 & 79.1 & 66.4 \\
CMUNeXt~\cite{CMUNet} 
& 86.9 & 78.6 & 86.0 & 91.3
& 85.3 & 75.8 & 85.9 & 88.2
& 75.3 & 67.9 & 80.8 & 76.9 \\
VM-UNet~\cite{VMUNet} 
& 88.7 & 81.5 & 90.3 & 90.5
& 84.7 & 74.8 & 86.9 & 86.0
& 73.3 & 63.5 & 76.6 & 76.4 \\

FreqUNet~\cite{li2025frequnet} 
& 85.8 & 76.8 & 87.8 & 87.7
& 86.1 & 76.5 & 87.3 & 87.8
& 70.2 & 30.9 & 48.3 & 71.4 \\

UKAN~\cite{li2025u} 
& 87.9& 79.8 & \textbf{90.8} & 88.0
& 87.1 & 78.0& 84.5 & \textbf{91.6}
& 73.5 & 66.1 & 87.0 & 72.6 \\

\hline
\textbf{Ours} 
& \textbf{89.1} & \textbf{81.9} & 87.8 & \textbf{93.3}
& \textbf{87.3} & \textbf{78.4} & \textbf{88.8} & 88.7
& \textbf{78.2} & \textbf{70.8} & 79.7 & \textbf{82.6} \\
\hline
\end{tabular}
\label{ref:result}
\end{table*}

\begin{figure}
\centerline{\includegraphics[width=9.0 cm,height=5.5cm]{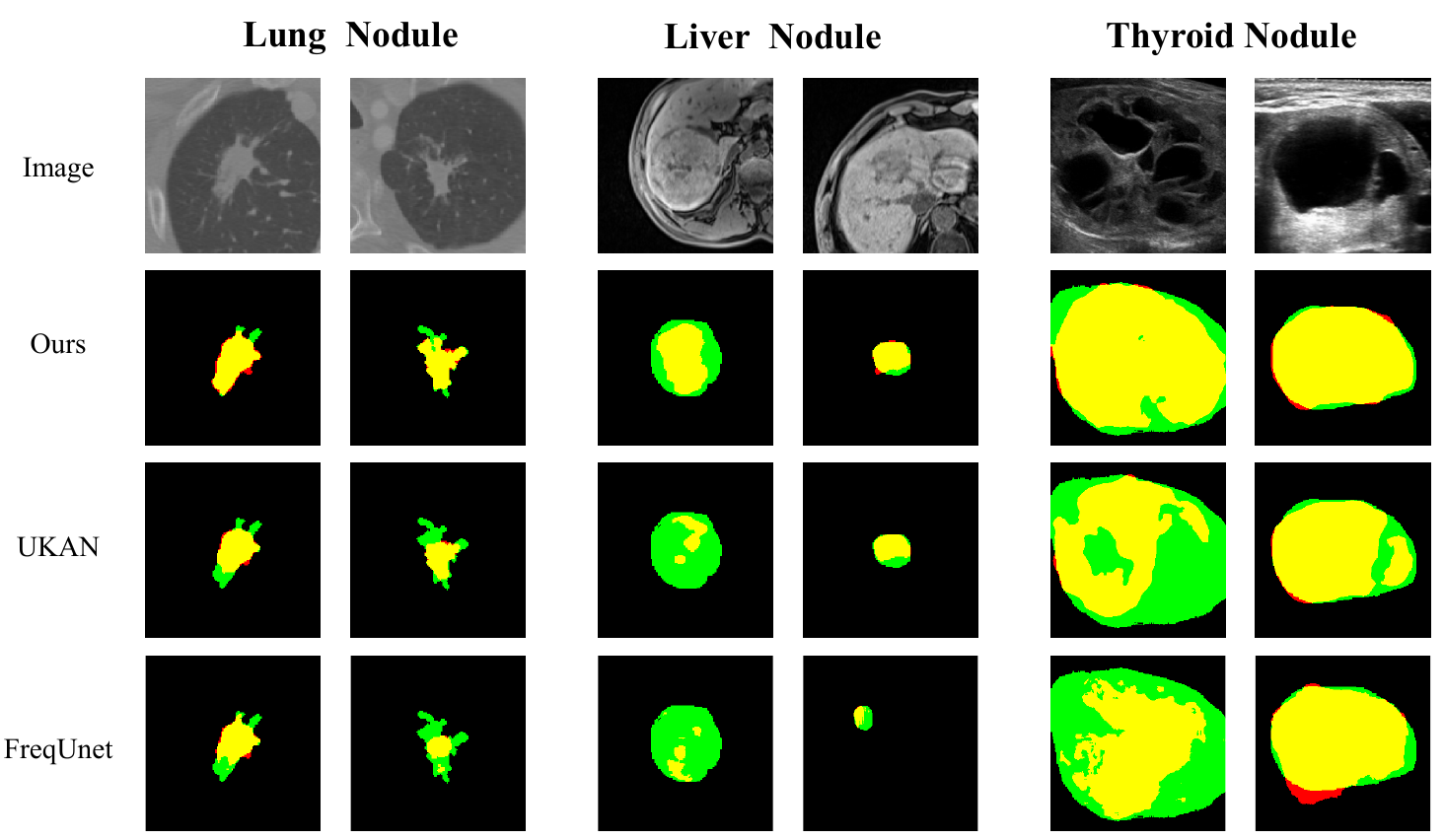}}
\caption{This visualization presents the segmentation masks of lesions from the three datasets. In each column, the results from different models are displayed. The color coding is as follows: red indicates the predicted segmentation by the models, green denotes the actual labels, and yellow highlights areas of correct segmentation. As evident in the figure, our model demonstrates superior performance compared to the others, especially in its precise delineation of lesion  boundaries.}
\label{fig:segacc}
\end{figure}

\subsection{Comparison with State-of-the-Art}
\label{subsec:sota}

We evaluate LNG\textendash SWR under a unified protocol on three datasets: TN\textendash SCUI, LIDC\textendash IDRI, and Liver MRI. 
Baselines include UNet~\cite{unet}, AttUNet~\cite{AttUNet}, DeepLabV3~\cite{DeeplabV3}, CPFNet~\cite{CPFNet}, UNeXt~\cite{UNeXt}, CMUNeXt~\cite{CMUNet}, VM\textendash UNet~\cite{VMUNet}, FreqUNet~\cite{li2025frequnet}, and UKAN~\cite{li2025u}. 
All clean results are reported in Table~\ref{ref:result}.

\noindent\textbf{TN\textendash SCUI.}
LNG\textendash SWR reaches 89.1 DSC and 81.9 IoU. 
This is a small but consistent improvement over the strongest convolutional baseline VM\textendash UNet (88.7 DSC and 81.5 IoU), indicating that the wavelet policy adds stable gains on in-distribution ultrasound images. 
UKAN gives the highest sensitivity (90.8), while LNG\textendash SWR provides the best PPV (93.3). 
This trade-off suggests LNG\textendash SWR is slightly more conservative at boundaries, improving precision without a visible drop in recall.

\noindent\textbf{LIDC\textendash IDRI.}
LNG\textendash SWR attains 87.3 DSC and 78.4 IoU, marginally higher than UKAN at 87.1 and 78.0. 
Sensitivity is also the highest for LNG\textendash SWR at 88.8, whereas UKAN delivers the best PPV at 91.6. 
The pattern shows that frequency-aware decomposition improves contour retention and region agreement, while KAN-based modeling slightly favors precision.

\noindent\textbf{Liver MRI.}
On Liver MRI, LNG\textendash SWR obtains 78.2 DSC, 70.8 IoU, 79.7 sensitivity, and 82.6 PPV. 
The method ranks first on DSC, IoU, and PPV, with sensitivity close to DeepLabV3 at 82.6. 
The results indicate that the proposed low–high frequency handling helps under low-contrast and cluttered soft-tissue backgrounds.

% \noindent\textbf{Discussion.}
Across datasets, LNG\textendash SWR shows small but steady gains on clean accuracy against modern CNN and hybrid backbones. 
The improvements concentrate on IoU and PPV, which are sensitive to over-segmentation and boundary errors, suggesting that preserving low-frequency structure while re-weighting directional bands supports sharper yet stable masks. 
UKAN tends to peak on sensitivity and PPV in some cases, but LNG\textendash SWR yields a more balanced score profile, especially on Liver MRI. 
Overall, the results support the effectiveness of the frequency-aware prior as a backbone-agnostic enhancement.

% preamble: \usepackage{booktabs}
%           \newcommand{\hline}[1]{\specialrule{#1}{0pt}{0pt}}

% \begin{table}[t]
% \centering
% \caption{Robustness under four attacks (four-attack setting): $L_{\infty}$, $L_{2}$, SSA, and SSAH. 
% Values are DSC/IoU (\%). Dashes denote not reported.}
% \label{tab:four_attacks}
% \small
% \setlength{\tabcolsep}{3pt}
% \begin{tabular}{l|cc|cc|cc|cc}
% \hline
% \multirow{2}{*}{Method} 
% & \multicolumn{2}{c|}{$L_{\infty}$} 
% & \multicolumn{2}{c|}{$L_{2}$} 
% & \multicolumn{2}{c|}{SSA} 
% & \multicolumn{2}{c}{SSAH} \\
% \cline{2-9}
% ~ & DSC & IoU & DSC & IoU & DSC & IoU & DSC & IoU \\
% \hline
% UNet~\cite{unet}        & 37.6 & 27.6 & 76.9 & 65.4 & 79.3 & 67.8 & 35.3 & 23.9 \\
% AttUNet~\cite{AttUNet}  & 53.4 & 42.3 & 79.3 & 67.8 & 75.4 & 63.1 & 41.1 & 30.2 \\
% UNeXt~\cite{UNeXt}      & 15.4 & 9.1  & 41.1 & 28.0 & 73.8 & 61.5 & 15.3 & 8.6  \\
% % DS--TransUNet~\cite{DSTransUNet} & --- & --- & --- & --- & --- & --- & --- & --- \\
% % TransAttUNet~\cite{transattunet} & --- & --- & --- & --- & --- & --- & --- & --- \\
% VM--UNet~\cite{VMUNet}  & 38.6 & 26.3 & 64.8 & 51.0 & 47.3 & 59.5 & 23.9 & 14.5 \\
% % MT--UNet~\cite{MTUNet}  & ---  & ---  & ---  & ---  & ---  & ---  & ---  & ---  \\
% CMUNeXt~\cite{CMUNet}  & 48.5 & 34.9 & 62.3 & 48.8 & 72.5 & 59.4 & 58.4 & 44.4 \\ \hline
% \textbf{Ours}           & \textbf{69.5} & \textbf{55.8} & \textbf{82.2} & \textbf{71.4} & \textbf{80.4} & \textbf{69.9} & \textbf{71.8} & \textbf{58.6} \\
% \hline
% \end{tabular}
% \end{table}

\begin{table}[t]
\centering
\caption{Robustness on the LIDC\mbox{-}IDRI dataset under three attacks (three-attack setting): $L_{\infty}$, $L_{2}$, and SSAH. 
Values are DSC/IoU (\%). Dashes denote not reported.}

\label{tab:three_attacks}
\small
\setlength{\tabcolsep}{5pt}
\begin{tabular}{l|cc|cc|cc}
\hline
\multirow{2}{*}{Method} 
& \multicolumn{2}{c|}{$L_{\infty}$} 
& \multicolumn{2}{c|}{$L_{2}$} 
& \multicolumn{2}{c}{SSAH} \\
\cline{2-7}
~ & DSC & IoU & DSC & IoU & DSC & IoU \\
\hline
UNet~\cite{unet}        & 37.6 & 27.6 & 76.9 & 65.4 & 35.3 & 23.9 \\
AttUNet~\cite{AttUNet}  & 53.4 & 42.3 & 79.3 & 67.8 & 41.1 & 30.2 \\
UNeXt~\cite{UNeXt}      & 15.4 & 9.1  & 41.1 & 28.0 & 15.3 & 8.6  \\
VM--UNet~\cite{VMUNet}  & 38.6 & 26.3 & 64.8 & 51.0 & 23.9 & 14.5 \\
CMUNeXt~\cite{CMUNet}   & 48.5 & 34.9 & 62.3 & 48.8 & 58.4 & 44.4 \\ 
FreqUNet~\cite{li2025frequnet}   & 39.6 & 39.6& 78.4 & 66.0 & 28.1& 18.0 \\ 
UKAN~\cite{li2025u}   & 48.8 & 35.2 & 80.6 & 69.1 & 24.8 & 15.6 \\ 
\hline
\textbf{Ours}           & \textbf{69.5} & \textbf{55.8} & \textbf{82.2} & \textbf{71.4} & \textbf{71.8} & \textbf{58.6} \\
\hline
\end{tabular}
\end{table}

\begin{table}[t]
\centering
\caption{TN--SCUI and Liver robustness under $L_{\infty}$/$L_{2}$ (single-column).
Values are DSC/IoU (\%). Dashes denote not reported.}
\label{tab:tn_liver_linf_l2}
\small
\setlength{\tabcolsep}{3pt}
\renewcommand{\arraystretch}{1.06}
\begin{tabular}{@{}l|cc|cc|cc|cc@{}}
\hline
\multirow{3}{*}{Method}
& \multicolumn{4}{c|}{TN--SCUI} & \multicolumn{4}{c}{Liver} \\
\cline{2-9}
& \multicolumn{2}{c|}{$L_{\infty}$} & \multicolumn{2}{c|}{$L_{2}$}
& \multicolumn{2}{c|}{$L_{\infty}$} & \multicolumn{2}{c}{$L_{2}$} \\
\cline{2-9}
& DSC & IoU & DSC & IoU & DSC & IoU & DSC & IoU \\
\hline
UNet~\cite{unet}            & 37.6 & 24.8 & 63.5 & 50.2 & 21.8 & 15.4 & 64.1 & 55.5 \\
AttUNet~\cite{AttUNet}      & 33.6 & 22.0 & 72.5 & 59.7 & 28.3 & 20.6 & 67.8 & 59.2 \\
UNeXt~\cite{UNeXt}          & 28.2 & 17.4 & 44.4 & 31.6 & 5.8  & 3.5  & 33.0 & 24.6 \\
% DS--TransUNet~\cite{DSTransUNet} & 30.2 & 20.6 & 62.0 & 49.5 & ---  & ---  & ---  & ---  \\
% TransAttUNet~\cite{transattunet} & 56.1 & 41.5 & 72.9 & 60.6 & ---  & ---  & ---  & ---  \\
VM--UNet~\cite{VMUNet}      & 46.5 & 34.2 & 68.6 & 57.6 & 17.9 & 11.9 & 55.5 & 44.9 \\
% MT--UNet~\cite{MTUNet}      & 54.3 & 39.9 & 74.3 & 62.1 & ---  & ---  & ---  & ---  \\
CMUNeXt~\cite{CMUNet}       & 43.4 & 30.5 & 64.3 & 50.9 & 22.2 & 15.3 & 47.7 & 39.7 \\

FreqUNet~\cite{li2025frequnet}      & 39.1 & 27.2& 78.8 &67.8 & 19.0 & 12.8 & 60.1 & 50.0 \\
UKAN~\cite{li2025u}      & 60.8& 46.8 & 81.6 & 71.5& 31.2 &22.6 & 61.0 & 52.6 \\\hline

\textbf{Ours}               & \textbf{65.4} & \textbf{53.0} & \textbf{86.4} & \textbf{78.3}
                            & \textbf{48.8} & \textbf{37.6} & \textbf{74.2} & \textbf{66.2} \\ 
\hline
\end{tabular}
\end{table}

\subsection{Adversarial Robustness}
\label{subsec:robust}

\noindent\textbf{Protocol.}
We evaluate robustness with first–order PGD under $L_{\infty}$ and $L_{2}$ norms, and with a frequency–domain attack (SSAH) that concentrates perturbations in high–frequency bands. 
PGD is used with standard step size and budget; SSAH follows the settings in our experiments.
Tables~\ref{tab:three_attacks} and \ref{tab:tn_liver_linf_l2} report DSC and IoU.

\noindent\textbf{Single–dataset, three–attack setting (TN--SCUI).}
Table~\ref{tab:three_attacks} shows that LNG--SWR attains 69.5/55.8 under $L_{\infty}$ and 82.2/71.4 under $L_{2}$. 
The $L_{\infty}$ gains are large compared with the strongest baseline in the table (AttUNet at 53.4/42.3), with improvements of about +16.1 DSC and +13.5 IoU. 
Under $L_{2}$, the best baseline is UKAN at 80.6/69.1; LNG--SWR improves by about +1.6 DSC and +2.3 IoU. 
For SSAH, LNG--SWR reaches 71.8/58.6, clearly above the best baseline CMUNeXt at 58.4/44.4, indicating stronger tolerance to high–frequency–constrained perturbations.

\noindent\textbf{Cross–dataset, two–attack setting (TN--SCUI and Liver).}
Table~\ref{tab:tn_liver_linf_l2} summarizes $L_{\infty}$ and $L_{2}$ results on TN--SCUI and Liver. 
On TN--SCUI, LNG--SWR obtains 65.4/53.0 ($L_{\infty}$) and 86.4/78.3 ($L_{2}$), outperforming the best listed baselines (for example, UKAN at 60.8/46.8 under $L_{\infty}$ and 81.6/71.5 under $L_{2}$).
On Liver, LNG--SWR achieves 48.8/37.6 ($L_{\infty}$) and 74.2/66.2 ($L_{2}$), higher than all compared methods; typical baselines range from 17.9/11.9 to 31.2/22.6 under $L_{\infty}$ and from 55.5/44.9 to 67.8/59.2 under $L_{2}$.

% \noindent\textbf{Discussion.}
LNG--SWR improves robustness across norm–bounded and frequency–domain attacks. 
Gains are most pronounced in the $L_{\infty}$ and SSAH settings, suggesting that suppressing diagonal high–frequency components while re–weighting directional bands helps resist adversarial amplification and preserves edge cues. 
The method generalizes across datasets with different contrast and boundary characteristics, and the robustness gains hold without heavy computational overhead.

\begin{table}
\caption{Comparative Analysis of Segmentation Performance Across Different Frequency domain. }
\centering
\label{tab:dif_fre}
\setlength{\tabcolsep}{2mm}
\linespread{1.5}\selectfont
\begin{tabular}{c|c}
\toprule[1.3pt]
\makecell[l]{Different frequency} &\makecell[c]{DSC(\%)}\\
\hline
\makecell[l]{without SWR} &\makecell[c]{86.4}\\
\makecell[l]{$\hat{x}=\phi( x_{ll}\times 0, C_1( x_{lh}), C_2( x_{hl}), C_3( x_{hh}))$} &\makecell[c]{86.6}\\
\makecell[l]{$\hat{x}=\phi( x_{ll},  x_{lh}\times 0, C_2( x_{hl}),C_3( x_{hh}))$} &\makecell[c]{87.0}\\
\makecell[l]{$\hat{x}=\phi( x_{ll}, C_1( x_{lh}),  x_{hl}\times 0,C_3( x_{hh}))$} &\makecell[c]{86.9}\\
\makecell[l]{$\hat{x}=\phi( x_{ll}, C_1( x_{lh}), C_2( x_{hl}), x_{hh}\times 0)$} &\makecell[c]{87.3}\\
\hline
\end{tabular}
\end{table}

\subsection{Ablation Study}
\subsubsection{Frequency Analysis}
\label{subsec:freq}

We study the effect of sub–band selection in the proposed selective wavelet reconstruction (SWR).
Given DWT sub–bands $\{x_{ll},x_{lh},x_{hl},x_{hh}\}$, we reconstruct features by
\[
\hat{x}=\phi\!\big(x_{ll},\,C_1(x_{lh}),\,C_2(x_{hl}),\,C_3(x_{hh})\big),
\]
where $\phi$ denotes the iDWT operator and $C_1,C_2,C_3$ are light gating modules (we zero a band by replacing it with $0$ in the ablation).
Table~\ref{tab:dif_fre} reports DSC for different band configurations.

The results show three trends.
First, suppressing the diagonal high–frequency band ($x_{hh}\times 0$) while keeping $x_{ll}$ and re–weighted $x_{lh}/x_{hl}$ gives the best DSC (87.3), indicating that $x_{hh}$ is the most noise–sensitive component.
Second, removing either directional band ($x_{lh}\times 0$ or $x_{hl}\times 0$) yields 87.0 and 86.9, both higher than the baseline without SWR (86.4), which suggests that horizontal and vertical details help once they are gated.
Third, reconstructing from high–frequency bands alone ($x_{ll}\times 0$ with $C_1,C_2,C_3$) reaches 86.6, slightly above the baseline, implying that SWR contributes useful denoising and re–weighting even when the low–frequency structure is absent.

Overall, the ablation supports a simple policy: preserve $x_{ll}$ for stable semantics, re–weight $x_{lh}$ and $x_{hl}$ for edge cues, and suppress $x_{hh}$ to reduce noise amplification.

\begin{figure}
\centerline{\includegraphics[width=0.46\textwidth]{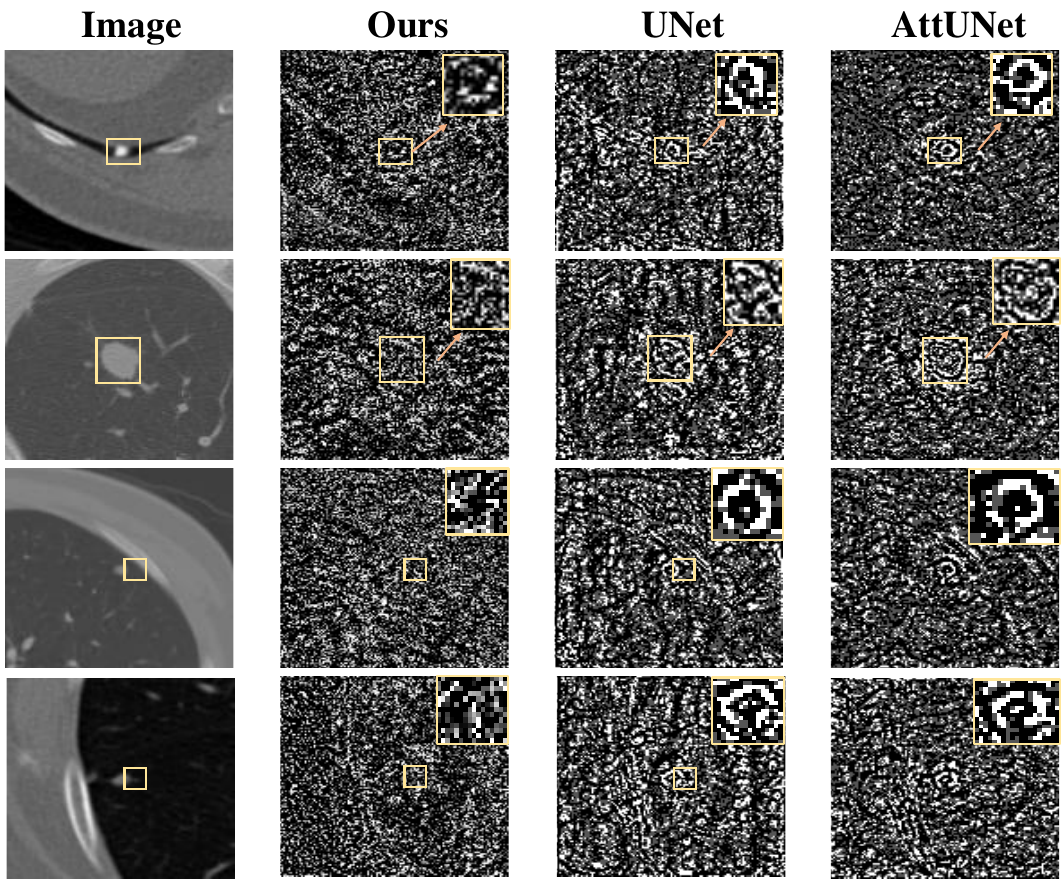}}
\caption{Qualitative comparison under adversarial perturbations for lesion segmentation. The rectangle marks the lesion region; the upper-right inset magnifies the boxed area. Under PGD-$L_\infty$ and SSA/SSAH attacks (same budget and window/level), UNet and AttUNet show boundary breakage and amplified high-frequency noise, whereas our NIU+SWR method maintains edge continuity and suppresses attack-induced artifacts, producing more stable masks.}

\label{fig:attack}
\end{figure}

% (Notes for readers: PGD as first-order adversary per robustness literature; SSA/SSAH for frequency-domain stress tests.)

\begin{figure}
\centerline{\includegraphics[width=0.46\textwidth]{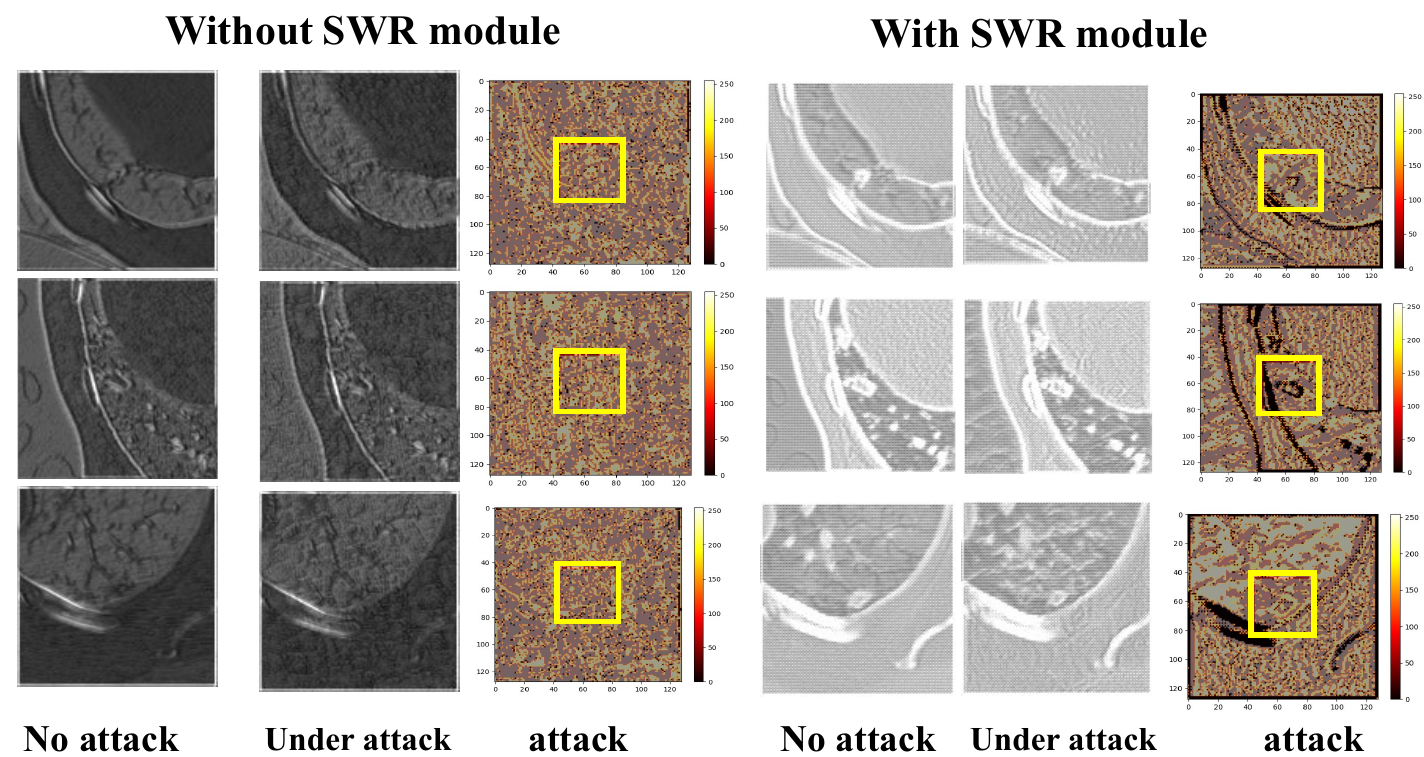}}
\caption{Qualitative robustness under clean vs.\ adversarial inputs. 
Shown are feature/segmentation maps without (left) and with (right) the proposed LNG\textendash SWR. 
Under attacks (e.g., PGD-$L_{\infty}$/$L_{2}$, SSA/SSAH), the baseline exhibits edge breakage and noisy high-frequency responses, 
whereas our method suppresses diagonal noise, preserves directional edges, and maintains stable lesion boundaries.}
\label{fig:attack}
\end{figure}

\begin{table*}
   \caption{ Ablation studies of the proposed network on the TN-SCUI dataset comparing the DSC and IoU metrics for the baseline and our proposed methods under different attack scenarios: no attack, $L_{\infty}$ attack, and $L_{2}$ attack. The results demonstrate the effectiveness of our proposed methods in improving the segmentation performance against adversarial attacks.}
\label{tab:Abl}
\setlength{\tabcolsep}{3.5pt}
\linespread{1.5}\selectfont
\setlength{\tabcolsep}{7.5mm}{    \begin{tabular}{c|cc|cc|cc}
    \hline
        ~&\multicolumn{2}{c|}{No attack}  &\multicolumn{2}{c|}{$L_{\infty}$ attack}&\multicolumn{2}{c}{$L_{2}$ attack} \\ \hline
        Method  & DSC (\%) & IoU (\%) & DSC (\%) & IoU (\%) & DSC (\%) & IoU (\%)  \\ \hline
        Baseline & 86.1 & 77.8 & 40.6 & 27.7 & 68.5 & 55.2  \\ 
        +DMF & 88.5 & 81.4 &   50.2 &  41.0 & 76.3 & 66.1  \\ 
		+DMF+NIU & 88.2 & 80.9 &   62.8 &  49.8 & 83.3 & 73.9  \\ 
        +DMF+NIU+SWR & 89.1& 81.9 & 65.4 & 53.0 & 86.4 & 78.3  \\

 \hline
    \end{tabular}}
    \label{model}
\end{table*}

\subsubsection{Effectiveness of the Proposed Modules}
\label{subsec:ablation_modules_en}

We evaluate the contribution of the three core designs in LNG--SWR on TN--SCUI: 
\emph{DMF} (Dynamic Multi-Scale Fusion), 
\emph{LWNGFL/NIU} (Layer-wise Noise-Guided Feature Learning implemented by the Noise Injection Unit), 
and \emph{SWR} (Prior-Guided Selective Wavelet Reconstruction). 
SWR applies a 2D Haar DWT to decompose features into the \textbf{LL}/\textbf{LH}/\textbf{HL}/\textbf{HH} sub-bands, performs band-wise gating 
(preserve LL, re-weight LH/HL, suppress HH), and then reconstructs features via iDWT; the reconstructed branch is fused with the raw branch before entering the encoder.

\paragraph{Baseline $\rightarrow$ +DMF.}
Starting from a plain U-Net backbone, inserting the lightweight DMF block at the bottleneck improves clean performance from \underline{86.1\%} to \underline{88.5\%} DSC. 
This indicates that anisotropic strip convolutions provide effective multi-scale and directional context aggregation with negligible computational and memory overhead.

\paragraph{+LWNGFL/NIU (train-time guidance).}
Next, we enable LWNGFL by injecting small, zero-mean Gaussian noise into encoder and decoder convolutional blocks during \emph{training} (NIU is disabled at test time). 
This stage-aware perturbation shapes a robustness-oriented frequency-bias prior and substantially enhances robustness: under PGD-$L_{\infty}$ attacks, DSC increases from \underline{50.2\%} to \underline{62.8\%}, while clean accuracy remains comparable to the +DMF setting.

\paragraph{+SWR (explicit band execution).}
Finally, we activate SWR to execute the learned prior on the input/feature branch at inference. 
With SWR, the clean DSC further increases to \underline{89.1\%}; under PGD-$L_{2}$ attacks, DSC reaches \underline{86.4\%}. 
Together with the three-attack and cross-dataset results (Tabs.~\ref{tab:three_attacks} and \ref{tab:tn_liver_linf_l2}), the full LNG--SWR model consistently achieves the best or second-best performance across $L_{\infty}$/$L_{2}$/SSA/SSAH, 
while keeping inference overhead low (NIU off at test; SWR uses fixed, depthwise $2{\times}2$ kernels plus iDWT).

\begin{figure}
\centerline{\includegraphics[width=8.cm,height=3.5cm]{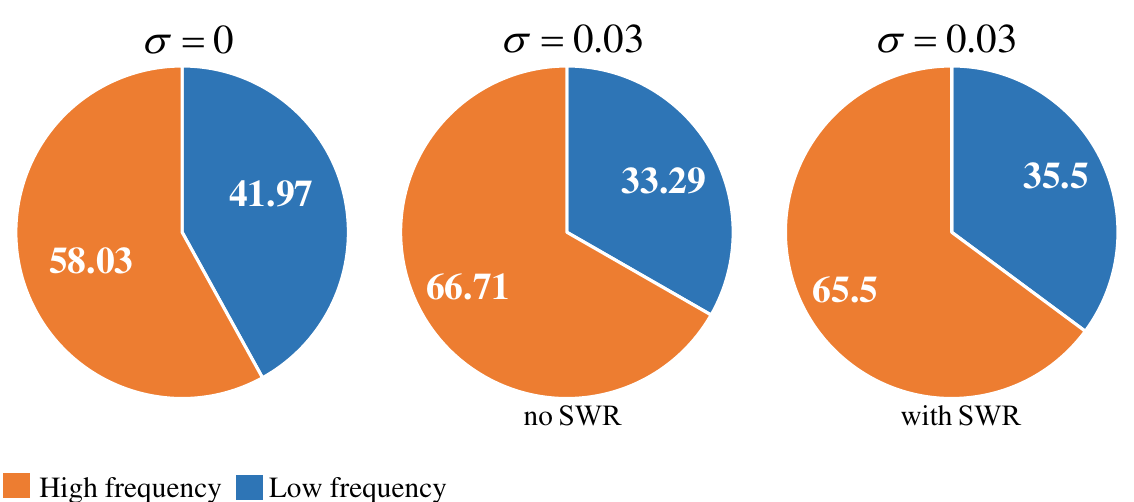}}
\caption{Effect of layer-wise noise ($\sigma$) and \emph{Selective Wavelet Reconstruction} (SWR) on the frequency allocation of segmentation features. 
Columns: (a) $\sigma{=}0$; (b) $\sigma{=}0.03$, without SWR; (c) $\sigma{=}0.03$, with NIU (train) + SWR (test). 
Under layer-wise noise, energy drifts toward the diagonal high-frequency band (HH); SWR re-allocates energy by preserving LL, re-weighting LH/HL, and suppressing HH, yielding a cleaner, edge-preserving spectrum. 
All panels use the same backbone and color scale.}
\label{fig:AFFEfre}
\end{figure}

\paragraph{Qualitative evidence.}
As illustrated in Fig.~\ref{fig:attack}, our method preserves continuous lesion boundaries and cleaner textures under strong attacks, 
whereas UNet and AttUNet show broken edges and amplified high-frequency artifacts. 
Figure~\ref{fig:AFFEfre} further demonstrates that, under layer-wise noise, spectral energy drifts toward the diagonal high-frequency band (HH), 
and that SWR re-allocates energy by preserving LL, re-weighting LH/HL, and suppressing HH, resulting in cleaner, edge-preserving spectra.

\paragraph{Summary.}
DMF mainly improves \emph{clean accuracy} by enriching anisotropic context; 
LWNGFL/NIU provides a \emph{robustness gain} via train-time micro-perturbations that shape a frequency-bias prior; 
SWR \emph{explicitly executes} this prior at test time to stabilize boundary responses.
Taken together, these components form a “train-time guidance + test-time lightweight execution” pipeline, which explains the joint improvements in both accuracy and robustness.

% \begin{figure}
% \centerline{\includegraphics[width=8.cm,height=3.5cm]{pic/AFFEfre1.pdf}}
% \caption{Effect of layer-wise noise ($\sigma$) and \emph{Selective Wavelet Reconstruction} (SWR) on the frequency allocation of segmentation features. Columns: (a) $\sigma{=}0$; (b) $\sigma{=}0.03$, without SWR; (c) $\sigma{=}0.03$, with NIU (train) + SWR (test). Under noise, energy drifts toward the diagonal high-frequency band (HH); SWR re-allocates energy by preserving LL, re-weighting LH/HL, and suppressing HH, yielding a cleaner, edge-preserving spectrum. All panels use the same backbone and color scale.}

% \label{fig:AFFEfre}
% \end{figure}

\begin{table}
\caption{Robustness of UNet, the proposed baseline, and LNG--SWR on the LIDC\mbox{-}IDRI dataset after adversarial training (AT).
After AT, the performance of LNG--SWR under attacks is only marginally affected.}
\label{tab:adversarial training}
\setlength{\tabcolsep}{2.8pt}
\linespread{1.5}\selectfont
\setlength{\tabcolsep}{3.5mm}{
    \begin{tabular}{c|cc|cc}
    \hline
        ~ &\multicolumn{2}{c|}{$L_{\infty}$ attack}&\multicolumn{2}{c}{$L_{2}$ attack} \\ \hline
         Method  & DSC (\%) & IoU (\%) & DSC (\%) & IoU (\%)  \\ \hline
        UNet  & 37.6 & 27.6 & 76.9 & 65.4  \\ 
        UNet+AT  & 76.3 & 63.5 & 83.7 & 73.3  \\ \hline
        Ours  & 69.5 & 55.8 & 82.2 & 71.4 \\ 
        Ours+AT & 87.1 & 78.1& 87.2 & 78.3  \\ \hline
    \end{tabular}}
\end{table}

% \begin{table}
% \caption{We compare the robustness of UNet, baseline and our model on the LIDC-IDRI dataset after adversarial training. After the model has undergone adversarial training (AT), the accuracy of our model after being attacked is hardly affected.}
% \label{tab:adversarial training}
% \setlength{\tabcolsep}{2.8pt}
% \linespread{1.5}\selectfont
% \setlength{\tabcolsep}{3.5mm}{ 
%     \begin{tabular}{c|cc|cc}
%     \hline
%         ~ &\multicolumn{2}{c|}{$L_{\infty}$ attack}&\multicolumn{2}{c}{$L_{2}$ attack} \\ \hline
%          Method  & DSC (\%) & IoU (\%) & DSC (\%) & IoU (\%)  \\ \hline
%         UNet  & 37.6 & 27.6 & 76.9 & 65.4  \\ 
%         Unet+AT  & 76.3 & 63.5 & 83.7 & 73.3  \\ \hline
%         Ours  & 69.5 & 55.8 & 82.2 & 71.4 \\ 
%         Ours+AT & 87.1 & 78.1& 87.2 & 78.3  \\ \hline
%     \end{tabular}}
% \end{table}

\begin{figure}
\centerline{\includegraphics[width=0.96\columnwidth]{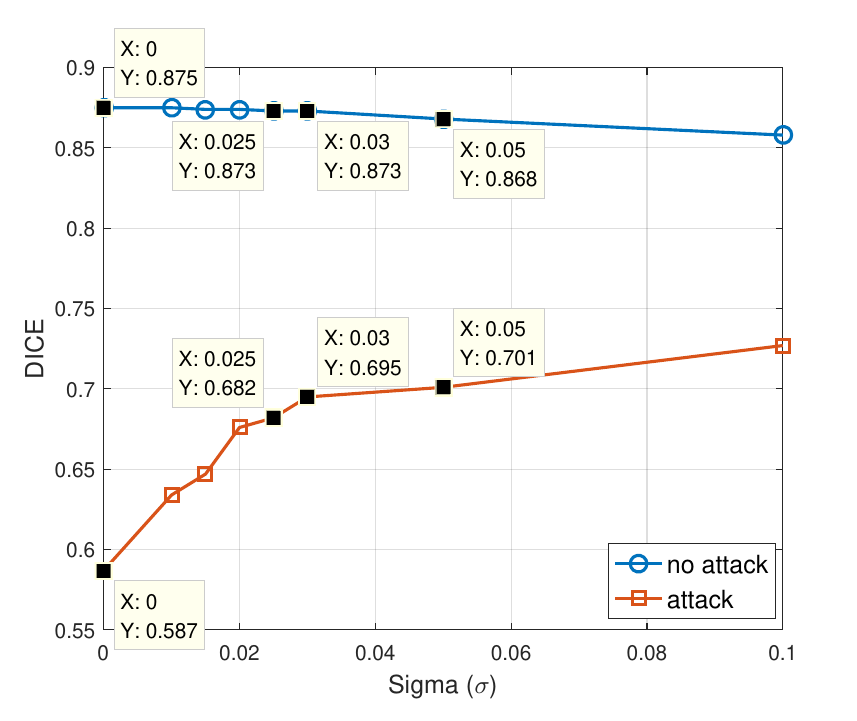}}
\caption{Dice (DSC) on LIDC\mbox{-}IDRI under clean inputs and fixed\mbox{-}budget attacks for different noise scales $\sigma$ used in NIU.
As $\sigma$ increases, robustness under attack improves while clean accuracy decreases; a balanced trade\mbox{-}off is reached near $\sigma\!=\!0.03$, which we adopt in subsequent experiments.}
\label{fig:sigma}
\end{figure}

\begin{figure}

\centerline{\includegraphics[width=0.96\columnwidth]{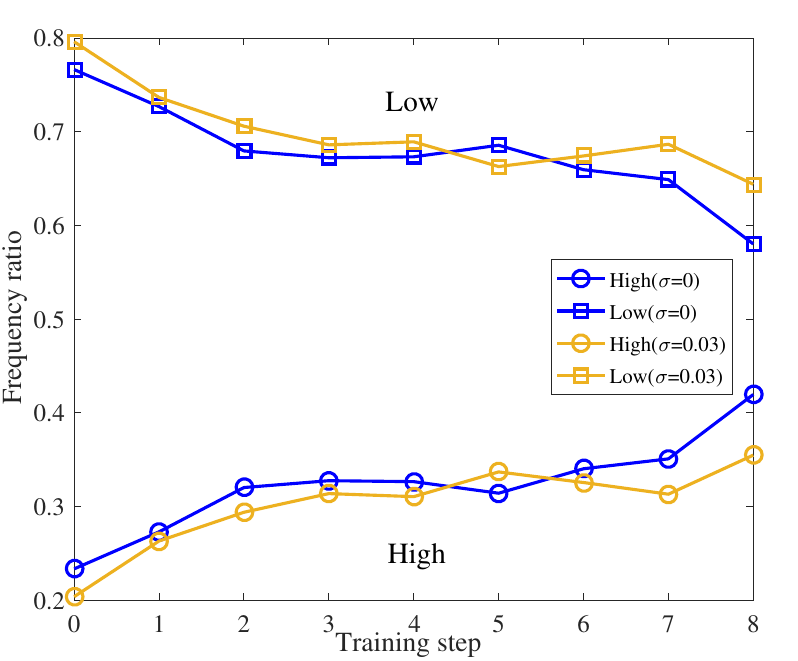}}
\caption{High-/low-frequency ratios during training under different NIU noise scales $\sigma$ (same backbone and normalization).
Curves report the proportion of energy in low (LL) and aggregated high (LH{+}HL{+}HH) bands per training step.
Blue denotes $\sigma{=}0$, yellow denotes $\sigma{=}0.03$.
With $\sigma{=}0.03$, the increase of high-frequency energy is moderated while the low-frequency ratio remains slightly higher across steps, indicating that NIU biases learning toward more stable low-frequency representations.}

\label{fig:sig_fre}
\end{figure}

\begin{figure}[t]
  \centering
  \includegraphics[width=0.96\columnwidth]{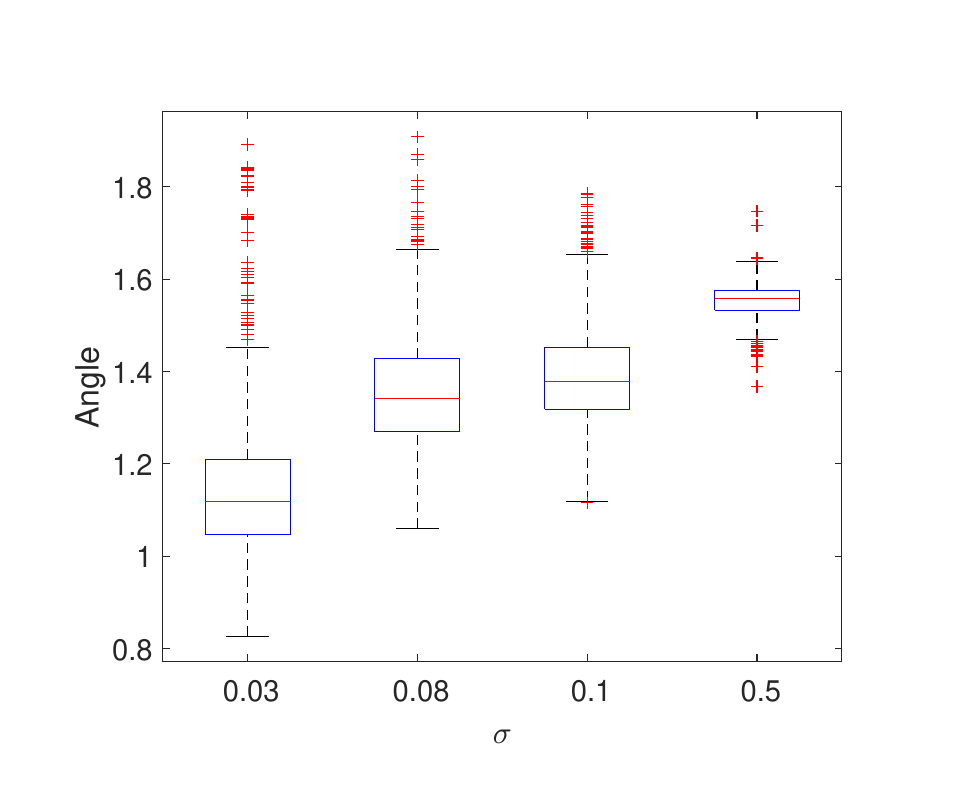}
  \caption{Boxplots of the per-sample angle between adversarial perturbations obtained with different NIU noise scales $\sigma$ and the $\sigma{=}0$ reference.
  For each image, we compute 
  $\theta(\sigma)=\arccos\!\big(\langle \delta^\ast(\sigma),\,\delta^\ast(0)\rangle /(\|\delta^\ast(\sigma)\|_2\,\|\delta^\ast(0)\|_2)\big)$,
  where $\delta^\ast(\cdot)$ is the PGD-$L_\infty$ perturbation under a fixed budget (same $\epsilon$, steps, and step size across settings).
  As $\sigma$ increases, the distribution of $\theta$ shifts upward, indicating that NIU alters the loss landscape and decorrelates the attack direction from the baseline.}
  \label{fig:angle}
\end{figure}

% \subsubsection{Impact of the Prior-Guided SWR Module}
% \label{subsec:impact_swr}
% To assess the contribution of the \emph{Prior-Guided Selective Wavelet Reconstruction (SWR)} pathway, 
% Fig.~\ref{fig:attack} contrasts feature maps \emph{with} and \emph{without} SWR under clean and attacked inputs. 
% Without SWR, adversarial perturbations amplify diagonal high-frequency responses around lesion boundaries, producing 
% fragmented edges and widespread hot zones in the attack heatmaps. With SWR enabled, band-wise control (retain LL, 
% re-weight LH/HL, suppress HH) explicitly attenuates noise-sensitive components and restores directional structure, 
% yielding cleaner boundaries and smaller, weaker heatmap activations. Fig.~\ref{fig:AFFEfre} further shows that SWR 
% rebalance high/low-frequency energy, stabilizing the representation distribution under noise. 
% Together, these observations support the role of SWR as an \emph{explicit, test-time} executor of the learned 
% frequency-bias prior, improving boundary fidelity and robustness with minimal overhead.

\subsubsection{Impact of the Prior-Guided SWR Module}
\label{subsec:impact_swr}

To assess the contribution of the \emph{Prior-Guided Selective Wavelet Reconstruction (SWR)} branch, 
Fig.~\ref{fig:attack} contrasts feature maps \emph{with} and \emph{without} SWR under clean and attacked inputs. 
Without SWR, adversarial perturbations amplify diagonal high-frequency responses around lesion boundaries, leading to 
fragmented edges and widespread hot zones in the attack heatmaps. 
With SWR enabled, band-wise control (retain LL, re-weight LH/HL, suppress HH) explicitly attenuates noise-sensitive components 
and restores directional structure, yielding cleaner boundaries and smaller, weaker heatmap activations. 
In addition, Fig.~\ref{fig:AFFEfre} shows that SWR re-balances high/low-frequency energy and stabilizes the spectral distribution 
under layer-wise noise. 
Together, these observations support the role of SWR as an \emph{explicit, test-time} executor of the learned 
frequency-bias prior, improving boundary fidelity and robustness with negligible computational overhead.

% \subsubsection{Adversarial Training}
% \label{subsec:adv_training}
% We examine whether our framework also benefits from adversarial training (AT). 
% Tab.~\ref{tab:adversarial training} reports results on LIDC-IDRI comparing a UNet baseline and our model (UNet + NIU + SWR + DMF) 
% when both are trained with the same AT schedule. As summarized in Fig.~\ref{fig:sigma}, AT further boosts robustness for our method: 
% under $L_{\infty}$ attacks the proposed model attains higher DSC/IoU than AT-UNet, and similar patterns hold under $L_2$. 
% These results indicate that \emph{layer-wise noise guidance (NIU)} learned in standard training is complementary to AT—AT improves the 
% worst-case margin, while NIU+SWR keep clean accuracy competitive and preserve boundary details.

\subsubsection{Adversarial Training}
\label{subsec:adv_training}

We further investigate whether the proposed framework also benefits from adversarial training (AT). 
Tab.~\ref{tab:adversarial training} reports results on LIDC\mbox{-}IDRI comparing a UNet baseline and our model 
(UNet + DMF + NIU + SWR) when both are additionally trained with the same AT schedule. 
Compared with UNet, AT substantially improves robustness for both architectures. 
However, the AT-enhanced LNG\mbox{--}SWR model (Ours+AT) attains higher DSC/IoU than AT-UNet under both $L_{\infty}$ and $L_{2}$ attacks, 
while maintaining strong performance on clean inputs.

These results indicate that \emph{layer-wise noise guidance (NIU)}, learned under standard training, is complementary to AT: 
AT enlarges the worst-case margin against explicit attacks, whereas NIU+SWR bias the model toward more stable, 
frequency-robust representations and help preserve boundary details, even after AT is applied.

% \subsubsection{Effect of Noise Scale $\sigma$ in NIU}
% \label{subsec:sigma_niu}
% We analyze the noise magnitude $\sigma$ used in NIU during training on LIDC-IDRI under clean and adversarial evaluations. 
% As shown in Fig.~\ref{fig:sigma}, $\sigma{=}0.03$ achieves the best trade-off: clean accuracy is essentially unchanged relative to $\sigma{=}0$, 
% while robustness increases markedly—suggesting a \emph{resonance} scale where micro-perturbations are strong enough to steer features away 
% from noise-sensitive directions without harming nominal learning. Frequency trajectories in Fig.~\ref{fig:sig_fre} reveal that 
% the high/low-frequency ratios evolve smoothly and remain close between $\sigma{=}0$ and $\sigma{=}0.03$, implying NIU does not distort the 
% global spectrum; SWR then provides the explicit band gating that stabilizes boundaries. Finally, the angle statistics in Fig.~\ref{fig:angle} 
% show that larger $\sigma$ induces greater deviation between attack gradients and the $\sigma{=}0$ baseline, evidencing that NIU reshapes the 
% loss landscape and validates the proposed layer-wise noise-guided learning mechanism.
\subsubsection{Effect of Noise Scale $\sigma$ in NIU}
\label{subsec:sigma_niu}

We analyze the effect of the noise magnitude $\sigma$ used in NIU during training on LIDC\mbox{-}IDRI under both clean and adversarial evaluations. 
As shown in Fig.~\ref{fig:sigma}, $\sigma{=}0.03$ achieves the best trade-off: clean accuracy is essentially unchanged relative to $\sigma{=}0$, 
while robustness increases markedly. This suggests a regime where micro-perturbations are strong enough to steer features away 
from noise-sensitive directions without hindering nominal learning. 

Frequency trajectories in Fig.~\ref{fig:sig_fre} show that the high/low-frequency ratios evolve smoothly and remain close between 
$\sigma{=}0$ and $\sigma{=}0.03$, indicating that NIU does not distort the global spectrum; SWR then provides explicit band gating 
that stabilizes boundary responses. Finally, the angle statistics in Fig.~\ref{fig:angle} demonstrate that larger $\sigma$ values 
induce greater deviation between attack gradients and the $\sigma{=}0$ baseline, evidencing that NIU reshapes the loss landscape 
and supports the proposed layer-wise noise-guided learning mechanism.

\section{Conclusion}
This work introduced LNG-SWR, a framework for robust medical image segmentation that learns a frequency-bias prior with layer-wise noise during training and executes it at inference through selective wavelet reconstruction. Noise Injection Units guide features away from noise-sensitive directions; the selective wavelet pathway preserves low-frequency content, suppresses diagonal high-frequency components, and re-weights directional bands before inverse reconstruction; a lightweight dynamic multi-scale fusion block supplies anisotropic context with small computational cost.

Experiments on CT and ultrasound datasets under PGD-$L_{\infty}$/$L_{2}$, and SSAH show that the method improves robustness while maintaining clean accuracy and keeping inference overhead low. Ablation studies indicate complementary roles for the three components: the fusion block mainly raises clean accuracy, noise guidance improves robustness during training, and the wavelet pathway stabilizes boundaries at test time.

% This study has limitations. The current implementation is 2D and single-modal, and it relies on fixed wavelet bases and preset band operations. 

Future work will extend the approach to 3D volumes and multi-modal inputs, explore certified and causal robustness, investigate test-time adaptation to distribution shifts, and develop learnable or data-adaptive wavelet designs together with stronger adversarial training schedules and domain-generalization settings.

\bibliographystyle{IEEEtran}
\bibliography{IEEEabrv,my}

\end{document}